\documentclass[a4paper,fleqn]{cas-dc}
\usepackage[numbers]{natbib}
\def\tsc#1{\csdef{#1}{\textsc{\lowercase{#1}}\xspace}}
\tsc{WGM}
\tsc{QE}
\usepackage{amsmath, amssymb, amsthm}
\usepackage{CJK, enumerate}

\usepackage[ruled,linesnumbered,vlined,algo2e]{algorithm2e}
\usepackage{rotating}
\usepackage{pdflscape}
\usepackage{url}
\usepackage{array}
\usepackage{breakurl}
\usepackage{stfloats}
\usepackage{bm}
\usepackage{algorithm}
\usepackage{algorithmic}
\usepackage{multirow}
\usepackage{makecell}
\usepackage[caption=false,,font=footnotesize]{subfig}

\def\etal{{\em et al.}}

\begin{document}
\begin{CJK*}{UTF8}{gkai}
\let\WriteBookmarks\relax
\def\floatpagepagefraction{1}
\def\textpagefraction{.001}

\shorttitle{LLM-SAEA for EOPs}    

\shortauthors{L. Xie \etal}

\title [mode = title]{Large Language Model-Driven Surrogate-Assisted Evolutionary Algorithm for Expensive Optimization}   

\author[a]{Lindong Xie}[orcid=0009-0004-2497-2338]

\ead{lindong.xie@connect.polyu.hk}

\author[b]{Genghui Li}[orcid=0000-0002-9950-9848]

\ead{genghuili2-c@my.cityu.edu.hk}

\author[c]{Zhenkun Wang}[orcid=0000-0003-1152-6780]
\ead{wangzhenkun90@gmail.com}

\author[a]{Edward Chung}[orcid=0000-0001-6969-7764]
\ead{edward.cs.chung@polyu.edu.hk}

\author[d,e]{Maoguo Gong}

\credit{Conceptualization, Methodology, Software, Experiments, Writing}

\affiliation[a]{organization={Department of Electrical and Electronic Engineering, The Hong Kong Polytechnic University},
            city={Hong Kong SAR},
            country={China}}

\affiliation[b]{organization={College of Computer Science and Software Engineering, Shenzhen University},
            city={Shenzhen},
            state={Guangdong},
            country={China}}

\affiliation[c]{organization={School of Automation and Intelligent Manufacturing, Southern University of Science and Technology},
            city={Shenzhen},
            state={Guangdong},
            country={China}}
            
            
\affiliation[d]{organization={Academy of Artificial Intelligence, College of Mathematics Science, Inner Mongolia Normal University},
            city={Hohhot},
            state={Inner Mongolia},
            country={China}}

\affiliation[e]{organization={Key Laboratory of Collaborative Intelligence Systems, Ministry of Education, Xidian University},
            city={Xi'an},
            state={Shaanxi},
            country={China}}

\credit{Writing, Review, Editing, Validation, Supervision}

\begin{abstract}
    Surrogate-assisted evolutionary algorithms (SAEAs) are a key tool for addressing costly optimization tasks, with their efficiency being heavily dependent on the selection of surrogate models and infill sampling criteria. However, designing an effective dynamic selection strategy for SAEAs is labor-intensive and requires substantial domain knowledge. To address this challenge, this paper proposes \textbf{LLM-SAEA}, a novel approach that integrates large language models (LLMs) to configure both surrogate models and infill sampling criteria online. Specifically, \textbf{LLM-SAEA} develops a collaboration-of-experts framework, where one LLM serves as a scoring expert (\textbf{LLM-SE}), assigning scores to surrogate models and infill sampling criteria based on their optimization performance, while another LLM acts as a decision expert (\textbf{LLM-DE}), selecting the appropriate configurations by analyzing their scores along with the current optimization state. Experimental results demonstrate that \textbf{LLM-SAEA} outperforms several state-of-the-art algorithms across standard test cases. The source code is publicly available at https://github.com/ForrestXie9/LLM-SAEA.
\end{abstract}






\maketitle

\section{Introduction}
A wide range of practical optimization tasks, including automotive crashworthiness analysis \cite{liu2021multisurrogate}, supply chain network design \cite{aldrighetti2021costs}, pharmaceutical drug discovery \cite{sadybekov2023computational}, and traffic network optimization \cite{ye2023initlight}, exhibit two primary characteristics. Firstly, they involve a black-box nature where the objective function lacks an explicit analytical expression \cite{jones1998efficient}. Secondly, solving these problems typically requires expensive simulations or time-consuming physical experiments \cite{lu2024you}. This category of problems is collectively known as expensive optimization problems (EOPs). Generally, an EOP can be mathematically represented as \cite{li2023evolutionary}:
\begin{equation}
\begin{split}
       & \text {min}\quad
       f(\textbf{x}),\\
       &\text {s.t. } \quad 
       \textbf{x} \in \mathbf{\Omega},
\end{split}
\label{eq:sop}
\end{equation}
where $\mathbf{x} \in \boldsymbol{\Omega} = [\mathbf{x}_l, \mathbf{x}_u] \subset \mathbb{R}^D$ is a $D$-dimensional decision vector bounded by $\mathbf{x}_l$ and $\mathbf{x}_u$, and $f(\mathbf{x})$ denotes the objective function mapping $\mathbf{x}$ to a scalar. In EOPs, $f(\mathbf{x})$ is assumed to be a black-box function, and its evaluation is computationally expensive \cite{he2023review}.

The surrogate-assisted evolutionary algorithm (SAEA) has become an effective and popular methodology for addressing EOPs \cite{wang2024customized, yan2025multiple}. It mainly comprises two core components: the surrogate model and the infill sampling criterion. The surrogate model is iteratively refined using evaluated solutions to approximate the black-box objective function. Based on the model's predictions, the infill sampling criterion recommends potentially promising solutions for costly evaluation in each iteration. Therefore, the efficiency of SAEA largely hinges on the selection of both the surrogate model and the infill sampling criterion. Depending on the selection mechanism used, previous SAEAs are generally categorized into two types: static configuration SAEAs \cite{li2025multi, tian2018multiobjective, li2020surrogate} and dynamic configuration SAEAs \cite{liu2022performance, zhen2022evolutionary}.

Static configuration SAEAs involve predetermining the surrogate model and infill sampling criteria before optimization begins and maintaining these configurations fixed throughout the entire optimization process \cite{xue2022multi}. A common approach is to fix a single surrogate model and pair it with a specific infill sampling criterion, such as a Gaussian process (GP) with lower confidence bound (LCB) \cite{liu2013gaussian} or expected improvement (EI) \cite{belakaria2020uncertainty}, a polynomial response surface (PRS) or radial basis function (RBF) with local search \cite{zhen2022evolutionary}, or k-nearest neighbors (KNN) with prescreening \cite{zhang2015multiobjective}. However, these straightforward combinations often struggle to effectively solve various optimization problems or perform consistently well across different optimization stages \cite{cai2019efficient}, due to the inherent biases of a single model and infill criterion \cite{wang2017committee}. Consequently, some prior works have integrated multiple models or infill sampling criteria to improve the generalization and robustness of SAEAs \cite{sonoda2022multiple, wu2022ensemble, zhai2024composite}. Although these methods aim to leverage the complementary strengths of different models and sampling criteria and typically outperform those using a single model or infill criterion in many cases, they also have some drawbacks. On the one hand, the simultaneous use of multiple models and/or infill criteria significantly increases computational complexity~\cite{shen2019meal}. On the other hand, some base models or sampling criteria may perform poorly in specific problem domains or optimization stages, thereby affecting overall optimization performance \cite{dong2020survey}.

To alleviate the limitations of static configuration SAEAs, dynamic configuration SAEAs have been proposed. These methods dynamically configure surrogate models and infill sampling criteria based on real-time feedback during the optimization process \cite{xie2023surrogate}. Typically, three strategies are developed for the automatic configuration of SAEAs: heuristic rules \cite{song2021kriging,liu2022performance}, reinforcement learning (RL) \cite{zhen2022evolutionary}, and multi-armed bandit (MAB) \cite{hoffman2011portfolio}. Heuristic rules, typically derived from the expertise of designers and grounded in well-established domain knowledge, are straightforward to design and implement but often struggle to adapt to dynamic and complex optimization environments due to their static nature \cite{iqbal2019actor,ma2024auto}. Moreover, the limited knowledge base of designers further restricts the generalizability of these rules. In contrast, RL and MAB utilize continuous interactions with the environment to optimize online configuration policy based on observed outcomes progressively. This adaptive capability significantly enhances the flexibility and efficacy of dynamic configuration in SAEAs. Despite these advancements, they still require labor-intensive, hand-crafted algorithmic components \cite{guo2024deep}, including the design of reward functions, the representation of actions and states, and the optimization of policies. These components are not only time-consuming to develop but also highly sensitive to task-specific settings. In light of these challenges, this work investigates the potential of large language models (LLMs) in assisting dynamic configuration within SAEAs. The motivation for exploring this shift stems from the inherent ability of LLMs to integrate diverse sources of domain knowledge (e.g., algorithm configuration and online learning), understand contextual information during the configuration process, and autonomously reason over configuration choices without the need for extensive manual engineering \cite{wu2024large}. Therefore, leveraging LLMs for dynamic configuration may serve as a viable and effective approach to automating algorithm configuration, thereby enhancing the efficacy of SAEAs in solving EOPs. Specifically, our contributions are summarized as follows:


\begin{itemize}
\item 
We develop a collaboration-of-experts framework that enables LLMs to automatically configure surrogate models and infill sampling criteria within the SAEA.

\item We utilize an LLM as a scoring expert, responsible for accurately assessing the utility of each selected model and its associated infill criterion.

\item We employ an LLM as a decision-making expert, responsible for recommending the suitable models and infill criteria at each iteration, based on their assessed utility and the optimization progress.

\item We validate that \textbf{LLM-SAEA} outperforms leading SAEAs on a range of benchmark problems. Additionally, we confirm the effectiveness of its algorithmic components through ablation experiments.

\end{itemize}

The paper is structured as follows: Section \uppercase\expandafter{\romannumeral2} provides a review of the relevant literature. Section \uppercase\expandafter{\romannumeral3} elaborates on the technical implementation details of the proposed algorithm \textbf{LLM-SAEA}. Section \uppercase\expandafter{\romannumeral4} presents a series of numerical experiments executed to verify the performance of the proposed algorithm. Section \uppercase\expandafter{\romannumeral5} offers concluding remarks and highlights potential avenues for future research.

\section{Related Work}
\subsection{Static Configuration SAEAs}
Existing related work on static configuration SAEAs can be broadly categorized into two types: static single-configuration SAEAs (SSC-SAEAs) and static multiple-configuration SAEAs (SMC-SAEAs). Among them, SSC-SAEAs are the most widely employed structure for SAEAs, requiring the selection of a surrogate model and an infill sampling criterion prior to the optimization commences. For example, GPEME \cite{liu2013gaussian} fixes the GP with LCB acquisition function to pre-screen high-quality solutions for true evaluation. Notably, when the decision variables exceed 50 dimensions, the GP is built in a lower-dimensional space obtained by the Sammon dimensionality reduction technique \cite{sammon1969nonlinear}. IKAEA \cite{zhan2021fast} first designs an incremental GP model to reduce the computational time of the GP. This model is subsequently combined with a standard EI acquisition function to predict promising solutions for real evaluation. CA-LLSO \cite{wei2020classifier} uses a small number of high-quality solutions to train a gradient-boosting classifier. Subsequently, an L1-exploitation sampling strategy is designed to leverage the level of information predicted by the classifier to further filter solutions for evaluation. Although SSC-SAEAs only require a single surrogate model paired with a specific infill sampling criterion, their rigidity can be a limitation in handling problems with diverse landscapes or varying requirements across iterations. To address this, SMC-SAEAs adapt hierarchical modeling strategies or integrate multiple models to enhance optimization performance. For example, ESAO \cite{wang2019novel}, GSGA \cite{cai2019efficient}, SA-MPSO \cite{liu2021surrogate}, and GL-SADE \cite{wang2022surrogate} implement a global-local modeling strategy to enhance the optimization process. The global model, constructed from all evaluated solutions, is aimed at exploring potentially promising regions by capturing broad trends in the search space. Conversely, the local model, trained on a subset of high-quality solutions, focuses on exploiting already identified promising areas by refining the search within those regions. HeE-MOEA \cite{guo2018heterogeneous} constructs an ensemble model that integrates an SVM and two RBF models, which are then coupled with the EI  and LCB acquisition functions to filter candidates for exact evaluations. ESCO \cite{wu2022ensemble} first trains multiple base models on low-dimensional data subsets. Then, a subset of these models with superior performance is selected to form a selective ensemble surrogate. Finally, the ensemble surrogate is combined with a specialized infill sampling strategy that accounts for both solution convergence and diversity to identify promising solutions for evaluation.

\subsection{Dynamic Configuration SAEAs}
Previous works have primarily used heuristic rules, RL, and MAB to accomplish the dynamic configuration of SAEAs. In the context of heuristic rules, existing approaches typically focus solely on configuring either the surrogate model or the infill sampling criteria. For instance, ASMEA~\cite{yu2020dynamic} initiates its process by establishing a surrogate pool, including radial basis functions, Gaussian process, linear Shepard interpolation, and polynomial response surfaces. Initially, some base models are filtered using two accuracy metrics (e.g., root mean square error (RMSE) and prediction residual error sum of squares). These filtered base models are then combined into five ensemble models and reintegrated into the model pool. Finally, the surrogate model with the lowest RMSE is dynamically chosen from the surrogate pool to guide the optimization process. Similarly, IBEA-MS~\cite{liu2022performance} adopts a reliability tolerance threshold to dynamically switch between the RBF and GP. In particular, when the predictive uncertainty associated with the GP model exceeds the specified threshold, the algorithm selects the RBF model; otherwise, the GP model continues to serve as the surrogate model. Regarding the heuristic rules for the adaptive selection of infill criteria, KTA2~\cite{song2021kriging} divides the optimization process into three distinct phases based on specific demands: convergence, diversity, and uncertainty. These phases are determined by calculating the distances between candidate solutions and the ideal reference point, alongside a diversity indicator. Leveraging this state estimation, an adaptive infill sampling strategy is developed to identify candidates for costly evaluations, optimizing the process according to the specific requirements of each phase. In contrast to KTA2, RVMM~\cite{liu2022reference} designs its adaptive sampling criteria based on only two optimization states: convergence and diversity. These two states are handled through two sets of reference vectors, with an adaptive set of reference vectors employed to promote convergence, whereas a fixed vector set is used to maintain solution diversity throughout the process. Regarding RL, 
ESA~\cite{zhen2022evolutionary} targets the adaptive selection of infill sampling strategies during optimization. It begins by assembling a diverse set of candidate strategies, such as DE-based evolutionary screening, full crossover, surrogate-assisted local exploration, and trust-region methods guided by surrogate models. It then employs Q-learning \cite{watkins1992q} to dynamically update the selection probabilities of these sampling strategies based on real-time feedback accumulated during the optimization. In the context of MAB, 
GP-Hedge~\cite{hoffman2011portfolio} employs an online MAB strategy to adaptively select an appropriate acquisition function based on cumulative rewards from a portfolio that includes widely used ones, including EI, probability of improvement, and upper confidence bound in each iteration. In contrast to GP-Hedge, AutoSAEA~\cite{xie2023surrogate} addresses the automated selection of surrogate models and sampling strategies by introducing a hierarchical MAB-based framework. This framework consists of two layers: the upper layer manages the selection of surrogate models, while the lower layer determines the infill sampling strategy. However, it requires careful manual design of both the reward function and selection strategy.

\subsection{LLMs for Automatic Algorithm Design}
In recent years, LLMs have garnered substantial attention and have been implemented across a diverse range of fields~\cite{floridi2020gpt,touvron2023llama,xu2024tad}. Some works have extended the capabilities of LLMs to automatic algorithm design, which can be categorized into two types: algorithm generation and algorithm selection. In terms of algorithm generation, a prevalent approach involves crafting specialized prompts that empower LLMs to directly facilitate optimization processes. For instance, LLAMBO~\cite{liu2024large} designs three prompts for LLMs specifically for warm-starting, sampling candidates, and surrogate modeling in Bayesian optimization (BO). CoE~\cite{xiao2023chain} and OptiMUS~\cite{ahmaditeshnizi2024optimus} interpret natural language descriptions of optimization problems, using LLMs to automate the development of mathematical models, write and debug solver code, and evaluate solutions. EvoLLM \cite{lange2024large} introduces a general prompt for LLMs in designing evolutionary strategies, which incorporates discretized solution candidate representations, performance-based least-to-most sorting, and fitness improvement query information. Another approach involves integrating LLMs with evolutionary algorithms. For example, FunSearch~\cite{romera2024mathematical}, AEL~\cite{liu2023algorithm}, EOH~\cite{liu2024evolution}, and LLaMEA~\cite {van2024llamea} integrate LLMs into evolutionary frameworks, employing operations such as mutation and/or crossover to enable the automated creation and refinement of algorithm code in each iteration. ReEvo~\cite{ye2024reevo} introduces short-term and long-term reflections within the evolutionary framework to efficiently explore the heuristic algorithm space. MEoH~\cite{yao2024multi} models heuristic algorithm search as a multi-objective optimization problem, with the algorithm's optimality and efficiency as the two objectives. Ultimately, it identifies a set of trade-off heuristic algorithms in a single run through a customized multi-objective search. Building on these contributions, several applications have further expanded upon these methods. For instance, FunBO~\cite{aglietti2024funbo} and EvolCAF~\cite{yao2024evolve} leverage these methods to generate and refine acquisition functions in BO, while L-AutoDA~\cite{guo2024autoda} efficiently designs competitive attack algorithms for image adversary attacks. Additionally, TS-EoH~\cite{yatong2024ts} develops heuristic algorithms for edge server task scheduling. Regarding algorithm selection, AS-LLM~\cite{wu2024large} extracts algorithm features using LLMs, which are then refined by a feature selection module. These features, combined with the problem representation, are input into a similarity calculation module to determine the best-matching algorithm.

\section{LLM-SAEA}
\subsection{Algorithm Structure}

\begin{algorithm}[!h]
    \begin{algorithmic}[1]
        \STATE \textbf{Input:}
        \STATE \quad Population size: $N$
        \STATE \quad Maximum number of FEs: $MFEs$

        \STATE \quad Combinatorial action set: 
        $\mathcal{CA}
        = \{a_1 ,\ldots , a_8\}$                
         \STATE \textbf{Initialization:}
        \STATE \quad Sample $N$ solutions using LHS

         \STATE \quad Evaluate and store them into the database $\mathcal{D}$ 
 
        \STATE \quad  Set  $FEs\leftarrow N$ and $t\leftarrow1$ 
        \STATE \quad Set $S_{a}(t) \leftarrow 0,  V_{a}(t) \leftarrow0,  \forall a \in \mathcal{CA}$
        
         \WHILE{ $FEs < MFEs $}
        \STATE  Select top $N$ solutions from $\mathcal{D}$ as population $\mathcal{P}$

        \STATE Get recommended action set $\mathcal{A}^{*}$ by \textbf{LLM-DE}
              \WHILE{$\mathcal{A}^{*} \neq \emptyset$} 

\STATE 
$a_t \gets \text{random.choice}(\mathcal{A}^{*})$
        
        \STATE 
        Obtain $\textbf{x}_t$ using $a_t$
        \STATE Evaluate $\textbf{x}_t$
        \STATE  $\mathcal{D} \leftarrow \mathcal{D}\cup  \textbf{x}_t$
        \STATE  $\mathcal{A}^{*} \leftarrow \mathcal{A}^{*} \setminus a_t$
        
        
        \STATE $FEs\gets FEs + 1$
        \STATE $T_{a_{t}}(t+1)\gets T_{a_{t}}(t)+1$
        \STATE Update the score and frequency of $a_t$ by \textbf{LLM-SE}         
        \STATE $t\gets t+1$
      
        \IF{$FEs \geq MFEs$ or $f(\textbf{x}_t)=\mathop{\mathrm{min}}_{\textbf{x} \in \mathcal{D}}f(\textbf{x})$}
        \STATE break

        \ENDIF
        
        \ENDWHILE  
        \ENDWHILE
 
        \STATE \textbf{Output:} The best solution in database $\mathcal{D}$
    \end{algorithmic}
\caption{\textbf{LLM-SAEA}($N$,  $MFEs$, 
$\mathcal{CA}$)}
\label{alg:LLM-SAEA}
\end{algorithm}

Algorithm \ref{alg:LLM-SAEA} illustrates the workflow of \textbf{LLM-SAEA}. Its inputs contain the population size $N$ (line 2), the maximum number of function evaluations $MFEs$ (line 3), and the combinatorial action set $\mathcal{CA}$ (line 4).
Each action in set \(\mathcal{CA}\) is a combination of a model and an infill sampling criterion. Specifically $\mathcal{CA}$ = \{(GP, LCB), (GP, EI), (RBF, Prescreening), (RBF, Local search), (PRS, Prescreening), (PRS, Local search), (KNN, L1-exploitation), (KNN, L1-exploration)\}. More details about these surrogate models and infill sampling criteria are provided in~\cite{he2023review,xie2023surrogate}.

In the initialization phase, $N$ initial solutions are sampled through Latin hypercube sampling (line 6). Subsequently, these solutions are assessed and stored in the database $\mathcal{D}$ (line 7). The number of function evaluations (FEs) is initialized to $N$, and the iteration counter $t$ is set to 1 (line 8). Additionally, the average score and selection frequency of each action in the combinatorial set $\mathcal{CA}$ is initialized to 0 (line 9). During the optimization process, the top $N$ solutions, based on fitness values, are selected from $\mathcal{D}$ to form the population $\mathcal{P}$ (line 11). Then, an LLM is employed as a decision expert (\textbf{LLM-DE}) to construct the candidate action set $\mathcal{A}^{*}$ (line 12). If $\mathcal{A}^{*}$ is not empty, an action $a_t$ is randomly chosen from $\mathcal{A}^{*}$ (lines 13-14). Specifically:
\begin{itemize}
\item If the selected action $a_t$ is one of \{(GP, LCB), (GP, EI), (RBF, Prescreening), (PRS, Prescreening), (KNN, L1-exploitation), (KNN, L1-exploration)\}, the differential evolution operator generates offspring set $\mathcal{X}=\{\textbf{o}_{1},\ldots,\textbf{o}_{N}\}$ based on the parent set $ \mathcal{P}$, as described in \cite{pant2020differential}:
\begin{equation}
\textbf{v}_{i}=\textbf{x}_{b}+F\times(\textbf{x}_{r1}-\textbf{x}_{r2}),
 \label{eq:de/current-to-best/1}
\end{equation}
\begin{equation}
       o_{i,j}=\left\{\begin{matrix}
         v_{i,j}& \mathrm{if} \ \ rand\leq CR \ \ \mathrm{or} \ \ j=j_{rand} \\
         x_{i,j}& \mathrm{otherwise}
     \end{matrix}\right.,
     \label{eq:bin}
\end{equation}
where the vector $\textbf{x}_{b}$ denotes the best solution in set $\mathcal{P}$, and $\textbf{x}_{r1}$ and $\textbf{x}_{r2}$ are randomly chosen from solution set $\mathcal{P}$. The scale factor $F$ is designated as 0.5. The values of $rand$ and $j_{rand}$ are independently sampled from the range $[0, 1]$ and the set $\{1, \dots, D\}$, respectively. The crossover rate $CR$ is fixed at 0.9. If the value of $o_{i,j}$ exceeds the predefined boundary constraints, it is subject to a corrective adjustment as follows:
\begin{equation}
    o_{i, j} = x_{l, j} + rand \times (x_{u, j} - x_{l, j}).
    \label{eq:repair}
\end{equation}

Subsequently, the chosen surrogate model (e.g., GP, RBF, or KNN) is first trained using the solution set $\mathcal{P}$ and then employed to estimate the the quality of the offspring in set $\mathcal{X}$, and its predictions are integrated into the selected infill sampling criterion to identify a promising solution $\textbf{x}_t$.


\item If the selected action $a_t$ is either (RBF, Local search) or (PRS, Local search), the chosen surrogate model (e.g., RBF or PRS) is trained on the set $\mathcal{P}$. Then, a differential evolution optimizer with a population size of $N$ and a total of $100D + 1000$ function evaluations~\cite{zhen2022evolutionary} is used to obtain a promising solution $\textbf{x}_t$ by optimizing the trained surrogate model $\hat{f}_{\mathrm{RBF/PRS}}$ within the subspace $\mathcal{X}$, as shown below:
\begin{equation}
\begin{aligned}
& \text {min} \quad \hat{f}_{\mathrm{RBF/PRS}}(\textbf{x}), \\
& \text {s.t.} \quad\ \ \textbf{x} \in \mathcal{X},
\end{aligned}
\end{equation}
where $ \mathcal{X} = [\textbf{lb}, \textbf{ub}]$ is the hyper-rectangular region; $\textbf{lb} = (l_1, \ldots, l_D)^{\top}$ and $\textbf{ub} = (u_1, \ldots, u_D)^{\top}$ denote the lower and upper bounds of the region, respectively; $l_j = \mathrm{min}\{x_{i,j} \mid \textbf{x}_{i} \in \mathcal{P}\}$ and $u_j = \mathrm{max}\{x_{i,j} \mid \textbf{x}_{i} \in \mathcal{P}\}$ for $j = 1, \ldots, D$.
\end{itemize}

\begin{figure}[ht]
        \centering
\includegraphics[height=1.65in,width=3.2in]{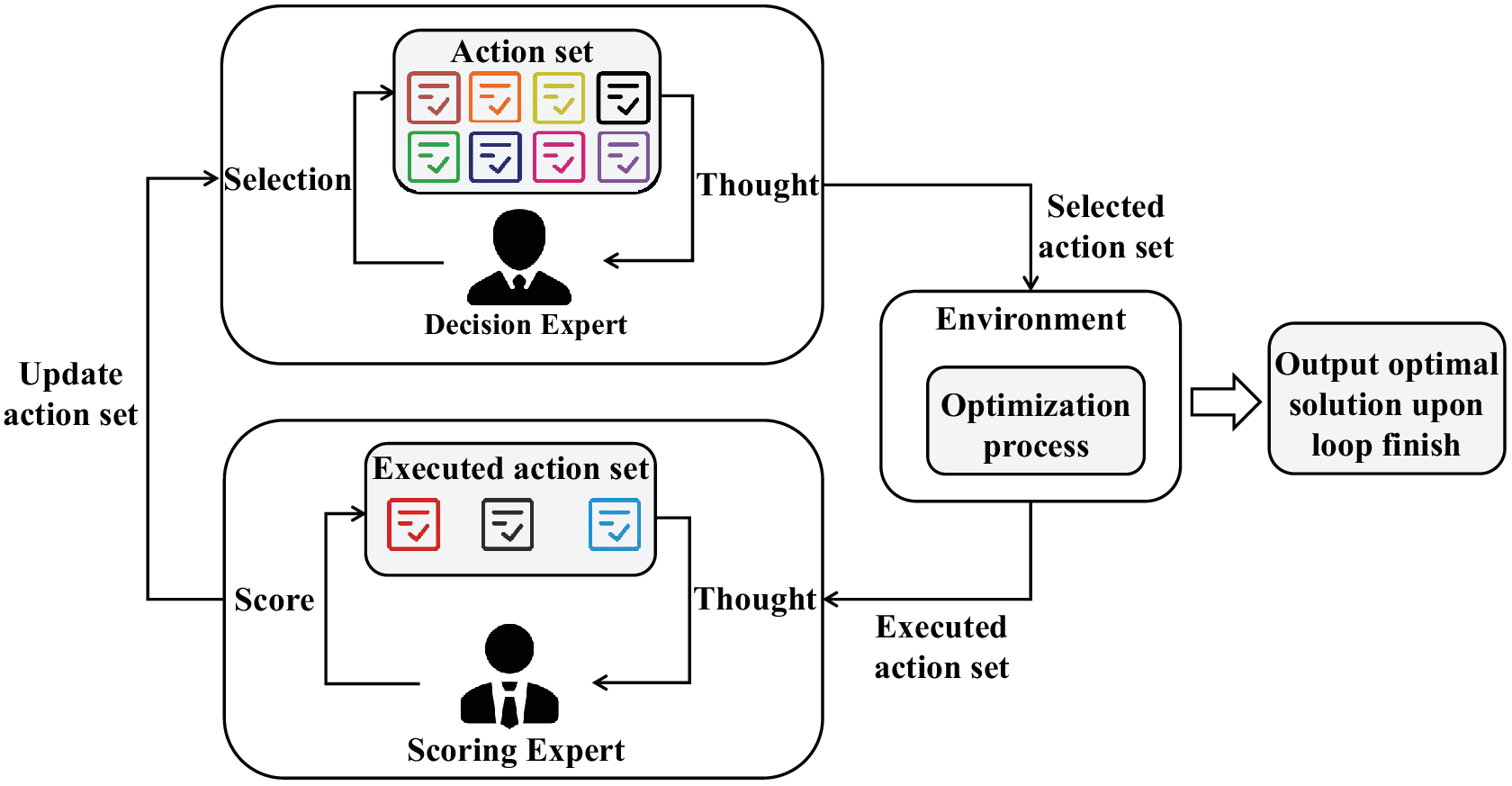}
        \caption{Illustration of the collaboration-of-experts framework. In each iteration, an LLM decision expert analyzes the contextual information of the action set to select actions. These selected actions are then randomly executed in the optimization environment until the current optimal solution is updated. An LLM scoring expert scores the executed actions and updates their contextual information. Once the stopping criterion is reached, the algorithm outputs the optimal solution found.
        }
        \label{fig: TEC}
\end{figure}

The solution $\textbf{x}_t$ is then evaluated and stored in the database $\mathcal{D}$ (lines 16-17). Following this, action ${a}_t$ is removed from the set $\mathcal{A}^{*}$ (line 18), and both the number of FEs and the selection frequency for action ${a}_t$ are incremented by 1 (lines 19-20). An LLM is employed as a scoring expert (\textbf{LLM-SE}) to update the average score and selection frequency of ${a}_t$ (line 21). It should be noted that if the termination condition is met or if the newly evaluated solution $\textbf{x}_t$ is the current best one in the database $\mathcal{D}$, the loop is exited (lines 23-25). Otherwise, actions in $\mathcal{A}^{*}$ are randomly selected one by one to provide new solutions for the expensive evaluation. Finally, when the stopping condition is reached, the optimal solution from the database $\mathcal{D}$ is output (line 28). To provide a clearer understanding of the workflow, a schematic diagram illustrating the process of \textbf{LLM-SAEA} is presented in Fig. \ref{fig: TEC}.

\subsection{Algorithm Core Components} 
\subsubsection{\textbf{LLM-DE}}
\begin{algorithm}[t]
    \begin{algorithmic}[1]
        \STATE \textbf{Input:}
        \STATE \quad Combinatorial action set: 
        $\mathcal{CA} = \{a_1, \ldots, a_8\}$ 
        \STATE \quad Total number of FEs: $MFEs$
         \STATE \quad Number of current FEs: $FEs$
        \STATE \quad 
         Average score of each action: $S(t)$
        \STATE \quad 
          Selected frequency of each action: $V(t)$
        \STATE \quad Number of iteration: $t$ 
        \STATE  Set $\mathcal{A}^{*} \gets \emptyset$ 

        \STATE  $(\mathcal{A}, \mathcal{L}) \gets \textbf{LLM}(\mathcal{CA}, S(t), V(t), MFEs, FEs, t)$

        \STATE  \textbf{for} $i\leftarrow 1$ \textbf{to} $|\mathcal{A}|$ \textbf{do}
        \STATE \quad \textbf{if} $\mathcal{L}\{i\} = \text{certain}$ \textbf{then}
        \STATE \quad \quad \textbf{if} $\mathcal{A}\{i\} \notin \mathcal{A}^{*}$ \textbf{then}
        \STATE  \quad \quad \quad $\mathcal{A}^{*} \leftarrow \mathcal{A}^{*} \cup \mathcal{A}\{i\}$
        \STATE \quad \quad \textbf{end if}
        \STATE  \quad \textbf{else}
        \STATE \quad \quad $P(t)\gets \text{Softmax}(S(t))$
        \STATE  \quad \quad  $a_r \gets \text{Roulette-wheel}(P(t))$ 

        \STATE  \quad \quad \textbf{if} $a_r \notin \mathcal{A}^{*}$ \textbf{then} 
        \STATE  \quad \quad \quad $\mathcal{A}^{*} \leftarrow \mathcal{A}^{*} \cup a_r$
        \STATE \quad \quad \textbf{end if}
        
        \STATE \quad \textbf{end if}
        \STATE  \textbf{end for}

        \STATE \textbf{Output:} Selected action set $\mathcal{A}^{*}$
        
    \end{algorithmic}

    \caption{\textbf{LLM-DE}($\mathcal{CA}$, $MFEs$, $FEs$, $S(t)$,$V(t)$, $t$)}
    \label{alg:LLM-DE}
\end{algorithm}

\begin{figure*}[!h]
        \centering
\includegraphics[height=2.7 in,width=6.3 in]{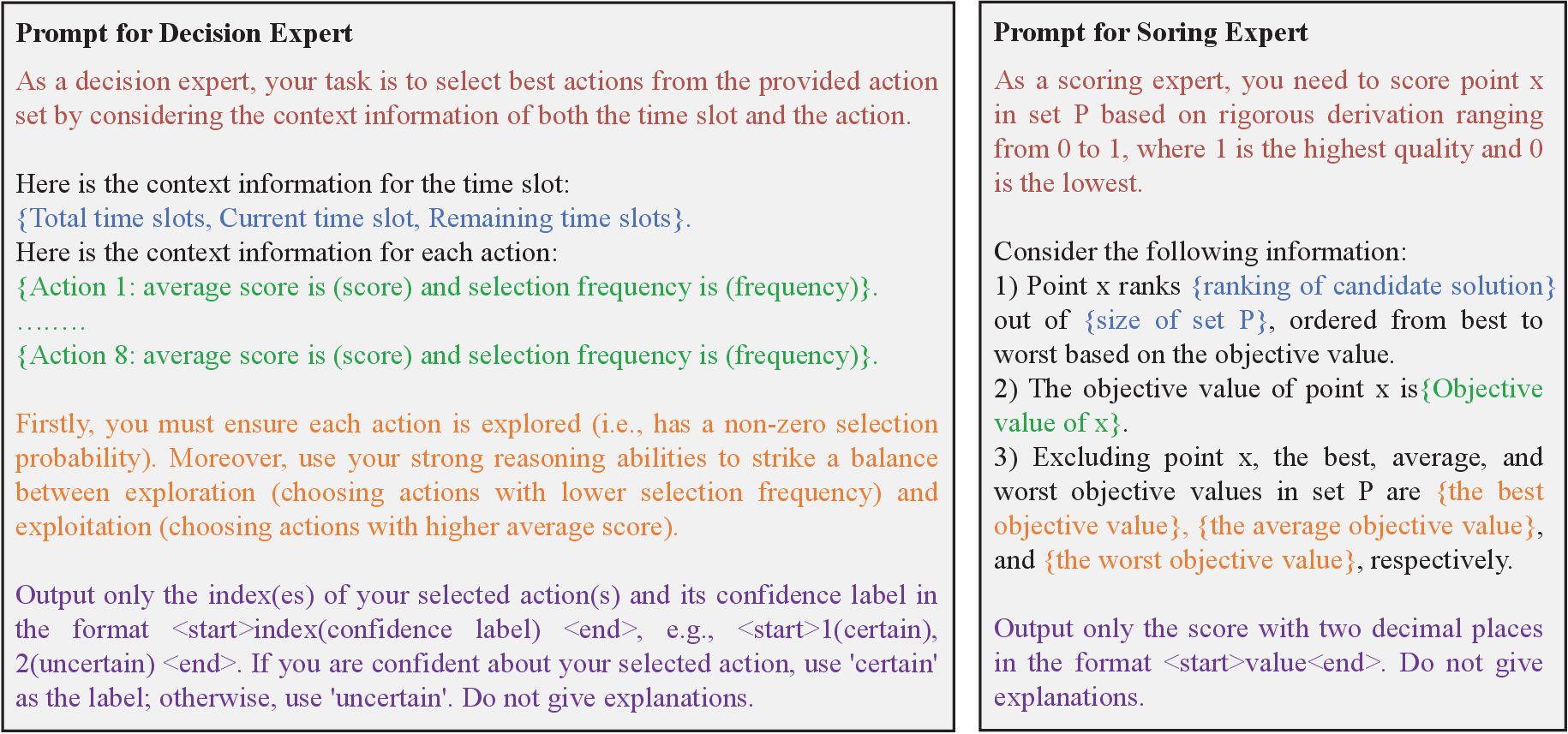}
        \caption{
        Prompts used in the decision expert and scoring expert. }
        \label{fig: Prompt}
\end{figure*}

The task of \textbf{LLM-DE} is to leverage the LLM as a decision expert for selecting actions. The corresponding pseudocode is provided in Algorithm~\ref{alg:LLM-DE}. Specifically, \textbf{LLM-DE} first observes the contextual information of the time slot and all actions in $\mathcal{CA}$, then outputs a selected set of actions $\mathcal{A}$ along with a corresponding 
set of confidence labels $\mathcal{L}$ for each selected action~(line 9). It should be noted that these confidence labels are generated via a self-reflection mechanism, wherein the LLM introspects and assesses the reliability of its own outputs. If the LLM is confident in its selected action, it is assigned the label 'certain'; otherwise, it receives the label 'uncertain'. The specific prompt for \textbf{LLM-DE} is shown in Fig. \ref{fig: Prompt}. Furthermore, if an action is labeled as 'certain', it is added to the set $\mathcal{A}^{*}$ (lines 11-14). Otherwise, the probabilities for each action are calculated using a softmax function based on their average scores (line 16), as described below: 
\begin{equation}
P_{a_i}(t) = \frac{\exp(S_{a_i}(t))}{\sum_{j=1}^{|\mathcal{CA}|} \exp(S_{a_j}(t))}, \quad i = 1, \ldots, |\mathcal{CA}|.
\end{equation}
Finally, the roulette wheel selection strategy is used to determine the final action $a_r$ according to the computed probabilities (line 17) , which is then added to $\mathcal{A}^{*}$ (lines 18-20).
\begin{algorithm}[t]
    \begin{algorithmic}[1]
        \STATE \textbf{Input:}
        \STATE \quad Population size: $N$
        \STATE \quad Population: $\mathcal{P}$
        \STATE \quad Newly evaluated solution: $\textbf{x}_{t}$  
        \STATE \quad Number of selections
        for $a_{t}$: $T_{a_{t}}(t)$ 
        \STATE \quad Number of iteration: $t$ 
        \STATE \quad 
        The average score for $a_t$: $S_{a_t}(t)$

        \STATE 
        $s_{a_t}\gets \textbf{LLM}(N, \mathcal{P}, \textbf{x}_t) $ 

        \STATE Update average score $S_{a_t}(t+1)$ for $a_t$ by (\ref{eq: update average score})

        \STATE Update selected frequency $V_{a_t}(t+1)$ of $a_t$ by (\ref{eq: update selected frequency})

        \STATE \textbf{Output:} The updated average score  $S_{a_t}(t+1)$ and selected frequency  $V_{a_t}(t+1)$.
    \end{algorithmic}
\caption{\textbf{LLM-SE}($N, \mathcal{P}, \textbf{x}_t, T_{a_t}(t), t, S_{a_t}$)}

\label{alg:LLM-SE}
\end{algorithm}
\subsubsection{\textbf{LLM-SE}}
After \textbf{LLM-DE} determines a set of actions $\mathcal{A}^{*}$, the actions within $\mathcal{A}^{*}$ have the opportunity to be executed during the optimization process. The executed action $a_t$ (model and infill sampling criterion) is employed to obtain the candidate solution $\textbf{x}_t$. Subsequently, \textbf{LLM-SE} is tasked with assessing the quality of $\textbf{x}_t$, as detailed in Algorithm \ref{alg:LLM-SE}. Specifically, \textbf{LLM-SE} employs the LLM as a scoring expert to score $\textbf{x}_t$ (line 8). The prompt for \textbf{LLM-SE} is shown in Fig. \ref{fig: Prompt}. The average score $S_{a_t}(t+1)$ and selection frequency $V_{a_t}(t+1)$ of action $a_t$ are updated in the following manner (lines 9-10):
\begin{equation}
         S_{a_t}(t+1) = \frac{T_{a_t}(t)S_{a_t}(t)+s_{a_t}}{T_{a_t}(t)+1},
         \label{eq: update average score}
     \end{equation}
\begin{equation}
         V_{a_t}(t+1) = \frac{T_{a_t}(t)+1}{t+1}.
         \label{eq: update selected frequency}
 \end{equation}
Finally, we present an analysis of the differences between the proposed \textbf{LLM-SAEA} and existing methods:
\begin{itemize}
\item  \textbf{LLM-SAEA} leverages the extensive prior knowledge and contextual understanding of LLMs to automate the selection of SAEA's core components (models and infill sampling criteria) through an online configuration way, thereby enhancing SAEA's performance and generalization capabilities.  

\item \textbf{LLM-SAEA} operates within a collaboration-of-experts framework where LLMs act as scoring and decision experts. They score and select actions based solely on specific task descriptions and relevant contextual information. This approach enhances the efficiency of algorithm configuration by eliminating the need for complex design work associated with dynamic algorithm configuration strategies.

\end{itemize}

\subsection{Algorithmic Complexity of \textbf{LLM-SAEA}}
The algorithmic complexity of the proposed \textbf{LLM-SAEA} is expressed as: 
\begin{equation} 
\mathcal{C}= MFEs \cdot C_{obj} + T \cdot \left[C_{\text{LLM-SE}} + C_{\text{LLM-DE}} + C_{SAEA}\right] \end{equation} 
where $MFEs \cdot C_{obj}$ denotes the cumulative computational cost associated with objective function evaluations over the maximum number of function evaluations (MFEs). The term $T$ refers to the number of optimization iterations and is defined as $T = MFEs - N$, where $N$ is the number of the initial population. The quantities $C_{\text{LLM-SE}}$ and $C_{\text{LLM-DE}}$ correspond to the computational costs incurred when utilizing the LLM as a decision expert and a scoring expert, respectively. $C_{SAEA}$ captures the cost of the surrogate-assisted evolutionary algorithm, which primarily comprises two components: the training of the surrogate model and the use of infill candidates based on predefined criteria. It is noteworthy that, in the context of expensive optimization problems, the term $MFEs \cdot C_{obj}$ typically constitutes the dominant contributor to the overall computational cost \cite{li2020surrogate}.

\section{Numerical Experiments}
\begin{table*}[t]
\renewcommand{\arraystretch}{1.1}
\tiny

\caption{The average and standard deviation of the function error values for all algorithms compared across the \textbf{10D} problems.
}
\centering
\resizebox{\textwidth}{!}{\begin{tabular}{p{0.7cm}p{0.1cm}p{1.85cm}p{1.85cm}p{1.85cm}p{1.85cm}p{1.85cm}p{1.85cm}p{1.85cm}p{1.85cm}p{1.85cm}p{1.85cm}}
\hline
\hline
Problem & D & ESAO & IKAEA & TS-DDEO & SA-MPSO & SAMFEO & GL-SADE & ESA & AutoSAEA & LLM-SAEA \\

Ellipsied & 10 & 7.55e-07(4.17e-07)+ & 8.75e-02(3.91e-01)+ & 5.28e-21(9.32e-21)+ & 6.66e-10(1.59e-09)+ & 2.67e-12(1.69e-12)+ & 1.06e-05(4.01e-06)+ & \cellcolor[rgb]{0.8,0.8,0.8}{\textbf{1.16e-34(3.27e-34)}}= & 3.06e-22(1.73e-22)+ & 6.25e-28(2.64e-27) \\
Rosenbrock & 10 & 4.66e-01(9.03e-01)= & \cellcolor[rgb]{0.8,0.8,0.8}{\textbf{5.15e-04(3.52e-04)}}- & 5.89e+00(1.88e-01)+ & 6.14e+00(5.31e-01)+ & 4.61e-01(4.73e-01)- & 8.63e+00(6.71e-01)+ & 1.76e+00(1.22e+00)+ & 2.54e+00(6.96e-01)+ & 6.27e-01(7.80e-01) \\
Ackley & 10 & 1.12e+01(3.30e+00)+ & 1.98e+00(1.63e+00)+ & 8.76e-04(1.16e-02)+ & 2.92e-04(7.75e-05)+ & 1.54e+00(1.44e+00)+ & 2.45e+00(1.86e+00)+ & 7.78e-01(9.58e-01)= & 2.56e-01(5.33e-01)- & \cellcolor[rgb]{0.8,0.8,0.8}{\textbf{1.15e-05(1.52e-05)}} \\
Griewank & 10 & 9.71e-01(8.20e-02)+ & 1.27e-01(1.96e-01)+ & 6.39e-01(1.55e-01)+ & 2.59e-02(1.58e-02)+ & 4.60e-02(4.10e-02)+ & 4.26e-02(3.97e-02)= & 3.25e-02(3.18e-02)+ & \cellcolor[rgb]{0.8,0.8,0.8}{\textbf{1.93e-03(3.91e-03)=}} & 1.65e-02(5.68e-02) \\
Rastrigin & 10 & 8.40e+01(1.35e+01)+ & 4.90e+01(2.29e+01)+ & 4.69e+01(9.05e+00)+ & 1.17e+01(4.76e+00)= & 2.33e+01(7.96e+00)+ & 2.45e+01(1.41e+01)= & 1.66e+01(1.03e+01)+ & 1.20e+01(3.97e+00)= & \cellcolor[rgb]{0.8,0.8,0.8}{\textbf{1.01e+01(6.34e+00)}} \\
F1 & 10 & 1.14e-05(5.87e-06)+ & 6.47e-02(9.53e-02)+ & 1.14e-13(4.52e-14)+ & 4.66e-07(1.84e-06)+ & 1.03e-10(5.07e-11)+ & 3.25e-06(8.04e-07)+ & \cellcolor[rgb]{0.8,0.8,0.8}{\textbf{0.00e+00(0.00e+00)}}= & \cellcolor[rgb]{0.8,0.8,0.8}{\textbf{0.00e+00(0.00e+00)}}= & \cellcolor[rgb]{0.8,0.8,0.8}{\textbf{0.00e+00(0.00e+00)}} \\
F2 & 10 & 2.73e+01(2.92e+01)+ & 7.23e-01(3.38e-01)+ & 2.16e+02(3.87e+02)+ & 3.97e+01(5.05e+01)+ & 3.27e-01(9.04e-01)+ & 6.01e+01(7.08e+01)+ & 1.34e-04(1.72e-04)+ & 6.04e-09(4.17e-09)+ & \cellcolor[rgb]{0.8,0.8,0.8}{\textbf{6.81e-10(9.35e-10)}} \\
F3 & 10 & 7.46e+05(4.49e+05)+ & 3.00e+05(1.73e+05)+ & 7.62e+06(6.95e+06)+ & 2.33e+06(2.10e+06)+ & 8.35e+05(9.46e+05)+ & 5.35e+05(8.47e+05)+ & 1.26e+05(1.19e+05)+ & 6.77e-02(5.38e-02)+ & \cellcolor[rgb]{0.8,0.8,0.8}{\textbf{5.13e-05(9.50e-05)}} \\
F4 & 10 & 6.62e+03(3.61e+03)+ & 1.62e+02(1.34e+02)+ & 1.20e+03(2.26e+03)+ & 5.79e+03(5.23e+03)+ & 2.96e+02(4.67e+02)+ & 7.11e+04(1.33e+04)+ & 1.36e+03(1.64e+03)+ & 6.69e+01(1.45e+02)+ & \cellcolor[rgb]{0.8,0.8,0.8}{\textbf{3.85e+00(1.19e+01)}} \\
F5 & 10 & 7.06e+02(5.01e+02)+ & 1.61e+03(1.29e+03)+ & 3.35e+02(1.20e+03)+ & 3.89e+00(4.59e+00)+ & 2.59e-02(2.36e-02)= & 5.81e+02(1.77e+03)+ & 1.24e+02(4.67e+01)+ & \cellcolor[rgb]{0.8,0.8,0.8}{\textbf{8.20e-03(2.40e-02)}}- & 4.32e-02(7.81e-02) \\
F6 & 10 & 2.54e+05(2.30e+05)+ & 2.51e+05(1.34e+05)+ & 1.66e+03(1.70e+03)+ & 1.96e+04(4.10e+04)+ & 1.13e+05(1.74e+05)+ & 1.50e+07(3.14e+07)+ & 3.68e+01(9.04e+01)+ & 6.36e+00(1.35e+00)= & \cellcolor[rgb]{0.8,0.8,0.8}{\textbf{6.04e+00(1.18e+00)}} \\
F7 & 10 & 1.27e+03(1.53e-02)+ & 1.32e+03(9.11e+01)+ & 1.27e+03(1.71e+00)+ & 1.27e+03(3.28e+00)+ & 1.27e+03(6.05e-02)- & 1.27e+03(6.80e-03)+ & 1.27e+03(1.88e+00)+ & \cellcolor[rgb]{0.8,0.8,0.8}{\textbf{1.27e+03(1.10e-04)}}- & 1.27e+03(2.00e-01) \\
F8 & 10 & 2.07e+01(1.18e-01)= & 2.08e+01(1.02e-01)+ & 2.07e+01(9.52e-02)= & 2.08e+01(1.29e-01)+ & 2.08e+01(1.34e-01)= & 2.07e+01(1.21e-01)= & 2.08e+01(1.33e-01)= & 2.07e+01(1.18e-01)= & \cellcolor[rgb]{0.8,0.8,0.8}{\textbf{2.07e+01(1.03e-01)}} \\
F9 & 10 & 7.88e+01(1.05e+01)+ & 6.52e+01(9.79e+00)+ & 4.64e+01(1.11e+01)+ & 1.55e+01(8.71e+00)= & 2.25e+01(9.85e+00)+ & 3.15e+01(1.10e+01)+ & 1.78e+01(8.50e+00)+ & 1.49e+01(6.30e+00)+ & \cellcolor[rgb]{0.8,0.8,0.8}{\textbf{1.03e+01(2.98e+00)}} \\
F10 & 10 & 7.88e+01(1.17e+01)+ & 7.67e+01(1.13e+01)+ & 5.85e+01(1.03e+01)+ & 1.92e+01(7.77e+00)+ & 2.32e+01(8.39e+00)+ & 2.34e+01(1.38e+01)+ & 2.00e+01(7.21e+00)+ & 1.26e+01(5.40e+00)= & \cellcolor[rgb]{0.8,0.8,0.8}{\textbf{1.20e+01(6.14e+00)}} \\
F11 & 10 & 1.20e+01(8.38e-01)+ & 5.58e+00(1.84e+00)- & 1.14e+02(1.18e+01)+ & 7.83e+00(1.80e+00)+ & 6.76e+00(2.19e+00)- & 8.42e+00(1.32e+00)+ & 6.19e+00(1.87e+00)= & \cellcolor[rgb]{0.8,0.8,0.8}{\textbf{5.44e+00(2.23e+00)}}- & 7.54e+00(2.24e+00) \\
F12 & 10 & 1.60e+03(4.47e+02)+ & 4.18e+03(8.88e+02)+ & 4.83e+02(7.88e+02)= & 1.47e+03(2.00e+03)= & 4.32e+03(8.62e+02)= & 5.60e+03(7.32e+03)+ & \cellcolor[rgb]{0.8,0.8,0.8}{\textbf{3.77e+02(6.58e+02)=}} & 3.97e+02(6.60e+02)= & 5.70e+02(7.16e+02) \\
F13 & 10 & 3.88e+01(3.71e+01)+ & 1.01e+02(7.32e+01)+ & 4.57e+00(1.05e+00)+ & 8.97e+00(2.27e+00)+ & 6.30e+00(1.59e+00)+ & 1.85e+03(1.27e+03)+ & 2.66e+00(1.07e+00)+ & 1.61e+00(1.07e+00)+ & \cellcolor[rgb]{0.8,0.8,0.8}{\textbf{1.32e+00(1.44e+00)}} \\
F14 & 10 & 4.40e+00(1.20e-01)+ & 3.96e+00(1.78e-01)= & 4.26e+00(1.59e-01)+ & 4.22e+00(1.52e-01)+ & 4.04e+00(2.16e-01)= & 4.24e+00(1.77e-01)+ & 3.98e+00(4.13e-01)= & \cellcolor[rgb]{0.8,0.8,0.8}{\textbf{3.88e+00(3.52e-01)=}} & 4.00e+00(1.99e-01) \\
F15 & 10 & 6.34e+02(3.68e+01)+ & 5.94e+02(2.91e+01)+ & 6.02e+02(7.30e+01)+ & 3.70e+02(1.11e+02)= & 3.91e+02(1.18e+02)= & 5.10e+02(1.27e+02)+ & 3.51e+02(1.178e+02)= & 3.74e+02(7.46e+01)+ & \cellcolor[rgb]{0.8,0.8,0.8}{\textbf{3.48e+02(1.21e+02)}} \\

\hline

+/=/-&& 18/2/0& 17/1/2  & 18/2/0  & 16/4/0  &12/5/3 &17/3/0&12/8/0 & 8/8/4  & NA\\
Ranking & &7.25 &6.20 & 5.95 & 5.50 &4.95&6.95&3.85&  2.25&\cellcolor[rgb]{0.8,0.8,0.8}{\textbf{{1.95 }}}\\

$p$-value   & &\cellcolor[rgb]{0.8,0.8,0.8}{\textbf{0.00}} &\cellcolor[rgb]{0.8,0.8,0.8}{\textbf{0.00}}&\cellcolor[rgb]{0.8,0.8,0.8}{\textbf{0.00}}&\cellcolor[rgb]{0.8,0.8,0.8}{\textbf{0.0002}}&\cellcolor[rgb]{0.8,0.8,0.8}{\textbf{0.002}}&\cellcolor[rgb]{0.8,0.8,0.8}{\textbf{0.00}}&\cellcolor[rgb]{0.8,0.8,0.8}{\textbf{0.05}}&0.73& NA\\

\hline
\hline
\end{tabular}}
\label{tab:10D_results}
\end{table*}
\subsection{Experimental Settings}
We evaluate the performance of \textbf{LLM-SAEA} using five widely recognized benchmark functions: Ellipsoid, Rosenbrock, Ackley, Griewank, and Rastrigin, along with 15 complex test problems (i.e., F1-F15 from~\cite{suganthan2005problem}) with dimensions $D=10$ and $D=30$. Each algorithm is evaluated for 1000 FEs and 20 independent runs per problem~\cite{liu2021surrogate, wang2022surrogate}. Moreover, we use a commonly employed \emph{function error metric} $f(\textbf{x}_{b}) - f(\textbf{x}_{o})$~\cite{li2020surrogate,xie2023surrogate} to quantify the performance of the algorithm, where $ \textbf{x}_{b}$ represents the optimum solution identified by the algorithm, and $ \textbf{x}_{o}$ denotes the true optimum of the test problem. Statistical significance of the results is analyzed using the Wilcoxon rank-sum test and the Friedman test with the Hommel post-hoc procedure at a significance level of 0.05 \cite{li2020surrogate} to demonstrate the differences in results among all tested algorithms. The symbols ``+'', ``='', and ``-'' indicate whether \textbf{LLM-SAEA} exhibits statistically superior, equivalent, or inferior performance compared to other algorithms, respectively.

\subsection{Comparison with State-of-the-art Algorithms}
\begin{table*}[b]
\renewcommand{\arraystretch}{1.1}
\tiny
\caption{The average and standard deviation of the function error values for all algorithms compared across the \textbf{30D} problems.
}
\centering
\resizebox{\textwidth}{!}{\begin{tabular}{p{0.7cm}p{0.1cm}p{1.85cm}p{1.85cm}p{1.85cm}p{1.85cm}p{1.85cm}p{1.85cm}p{1.85cm}p{1.85cm}p{1.85cm}p{1.85cm}}
\hline
\hline
Problem & D & ESAO & IKAEA & TS-DDEO & SA-MPSO & SAMFEO & GL-SADE & ESA & AutoSAEA & LLM-SAEA \\

       Ellipsied & 30 & 1.03e-04(3.40e-05)+ & 2.93e-03(1.08e-03) + & 1.18e-13(1.83e-13)+ & 7.76e-08(4.69e-08)+ & 1.10e-10(1.80e-10)+ & 3.12e-04(2.29e-04)+ & \cellcolor[rgb]{0.8,0.8,0.8}{\textbf{2.21e-22(4.40e-22)}}- & 2.21e-09(7.65e-10)+ & 1.27e-17(1.53e-17) \\
        Rosenbrock & 30 & 2.66e+01(1.18e+01)= & 6.13e+01(2.52e+01)+ & 2.71e+01(2.93e-01)+ & 4.24e+01(2.08e+01)+ & 2.74e+01(1.50e+00)+ & 3.00e+01(1.69e+00)+ & 2.65e+01(8.22e-01)= & 2.60e+01(1.31e+00)= & \cellcolor[rgb]{0.8,0.8,0.8}{\textbf{2.51e+01(7.27e-01)}} \\
        Ackley & 30 & 3.97e+00(5.92e-01)+ & 4.40e+00(6.09e-01)+ & 1.36e-04(6.21e-05)= & 7.77e-03(1.60e-02)+ & 2.51e+00(6.93e-01)= & 2.46e+00(5.72e-01)+ & \cellcolor[rgb]{0.8,0.8,0.8}{\textbf{8.11e-05(3.82e-05)}}- & 1.51e-01(3.71e-01)- & 1.70e+00(3.65e+00) \\
        Griewank & 30 & 9.51e-01(3.61e-02)+ & 9.62e-01(2.40e-02)+ & 3.31e-03(5.52e-03)+ & 2.04e-02(3.50e-02)+ & 4.26e-03(6.09e-03)+ & 1.71e-03(4.30e-03)+ & 1.72e-03(3.21e-03)+ & 1.91e-03(4.23e-03)+ & \cellcolor[rgb]{0.8,0.8,0.8}{\textbf{1.02e-03(7.00e-03)}} \\
        Rastrigin & 30 & 2.38e+02(2.68e+01)+ & 2.46e+02(2.31e+01)+ & 2.09e+02(1.44e+01)+ & 6.90e+01(1.56e+01)+ & 9.18e+01(2.93e+01)+ & 9.45e+01(3.14e+01)+ & 5.53e+01(4.94e+01)= & 4.76e+01(1.94e+01)= & \cellcolor[rgb]{0.8,0.8,0.8}{\textbf{4.06e+01(6.84e+01)}} \\
        F1 & 30 & 2.66e-04(5.18e-05)+ & 4.47e-02(2.48e-02)+ & 3.92e-12(4.33e-12)+ & 3.58e-04(1.27e+00)+ & 7.73e-10(1.17e-09)+ & 1.84e+02(2.40e+02)+ & \cellcolor[rgb]{0.8,0.8,0.8}{\textbf{1.14e-13(4.88e-14)}}- & 9.39e-10(2.27e-10)+ & 1.02e-13(2.73e-14) \\
        F2 & 30 & 4.83e+04(9.90e+03)+ & 6.24e+04(1.26e+04)+ & 3.34e+04(8.25e+03)+ & 4.21e+04(8.46e+03)+ & 1.89e+04(4.96e+03)+ & 5.37e+08(1.19e+04)+ & 2.23e+04(4.96e+03)+ & 1.30e+04(5.32e+03)= & \cellcolor[rgb]{0.8,0.8,0.8}{\textbf{1.21e+04(2.78e+03)}} \\
        F3 & 30 & 1.10e+08(5.07e+07)+ & 5.96e+07(1.88e+07)+ & 5.32e+07(2.64e+07)+ & 6.42e+07(3.17e+07)+ & \cellcolor[rgb]{0.8,0.8,0.8}{\textbf{1.06e+07(4.66e+06)}}= & 2.81e+07(1.26e+07)+ & 3.35e+07(1.73e+07)+ & 1.72e+07(9.05e+06)= & 1.45e+07(5.69e+06) \\
        F4 & 30 & 1.00e+05(2.54e+04)+ & 7.67e+04(1.64e+04)+ & 5.12e+04(1.43e+04)+ & 6.02e+04(1.58e+04)+ & 3.97e+04(9.12e+03)= & 7.38e+04(1.33e+04)+ & 3.79e+04(9.14e+03)= & 3.77e+04(7.92e+03)= & \cellcolor[rgb]{0.8,0.8,0.8}{\textbf{3.77e+04(6.64e+03)}} \\
        F5 & 30 & 2.52e+04(1.17e+03)+ & 9.71e+03(1.52e+03)+ & 6.63e+03(1.45e+03)+ & 9.17e+03(1.57e+03)+ & 8.20e+03(2.06e+03)+ & 8.85e+03(2.70e+03)+ & 5.98e+03(1.53e+03)= & 6.45e+03(1.19e+03)= & \cellcolor[rgb]{0.8,0.8,0.8}{\textbf{5.87e+03(1.49e+03)}} \\
        F6 & 30 & 2.65e+04(2.90e+04)+ & 1.43e+08(6.61e+07)+ & 2.54e+03(3.17e+03)= & 1.28e+07(7.92e+06)+ & 3.07e+06(2.06e+06)+ & 1.06e+08(6.33e+07)+ & 2.52e+03(2.57e+03)= & \cellcolor[rgb]{0.8,0.8,0.8}{\textbf{1.45e+03(2.19e+03)}}= & 1.85e+03(2.68e+03) \\
        
F7 & 30 & 4.70e+03(5.13e-07)- & 5.22e+03(2.76e+02)+ & 4.71e+03(3.34e+00)- & 5.51e+03(1.46e+02)+ & \cellcolor[rgb]{0.8,0.8,0.8}{\textbf{4.70e+03(5.07e-09)}}- & 4.70e+03(1.70e+01)- & 4.74e+03(1.06e+01)= & 4.70e+03(7.47e-01)- & 4.75e+03(4.20e+01) \\
F8 & 30& 2.12e+01(7.48e-02)= & 2.12e+01(7.28e-02)= & 2.12e+01(5.71e-02)= & 2.12e+01(6.71e-02)= & 2.12e+01(8.15e-02)= & 2.12e+01(7.35e-02)= & 2.12e+01(6.68e-02)= & 2.12e+01(6.71e-02)= & \cellcolor[rgb]{0.8,0.8,0.8}{\textbf{2.12e+01(4.76e-02)}} \\
F9 & 30& 2.40e+02(2.27e+01)+ & 2.53e+02(2.12e+01)+ & 2.13e+02(2.44e+01)+ & 8.94e+01(3.87e+01)+ & 7.93e+01(1.60e+01)= & 1.18e+02(2.61e+01)+ & 1.71e+02(4.29e+01)+ & \cellcolor[rgb]{0.8,0.8,0.8}{\textbf{4.86e+01(1.17e+01)}}= & 6.91e+01(4.88e+01) \\
F10 & 30& 2.94e+02(2.65e+01)+ & 3.05e+02(2.05e+01)+ & 2.32e+02(1.53e+01)+ & 1.44e+02(4.37e+01)+ & 1.01e+02(2.79e+01)+ & 1.32e+02(3.66e+01)+ & 2.20e+02(3.18e+01)+ & \cellcolor[rgb]{0.8,0.8,0.8}{\textbf{6.64e+01(1.49e+01)}}= & 7.57e+01(2.38e+01) \\
F11 & 30& 4.60e+01(1.75e+00)+ & 4.28e+01(5.25e+00)+ & 4.58e+01(1.24e+00)+ & 3.77e+01(3.80e+00)= & \cellcolor[rgb]{0.8,0.8,0.8}{\textbf{3.01e+01(4.28e+00)}}- & 3.10e+01(5.23e+00)- & 3.13e+01(7.37e+00)= & 4.25e+01(6.72e+00)+ & 3.80e+01(8.01e+01) \\
F12 & 30& 3.74e+04(3.94e+04)= & 4.71e+04(3.26e+04)+ & 6.92e+04(4.43e+04)+ & 1.33e+05(7.55e+04)+ & 4.58e+04(2.66e+04)+ & 8.65e+04(4.79e+04)+ & 5.30e+04(6.82e+04)= & \cellcolor[rgb]{0.8,0.8,0.8}{\textbf{1.59e+04(1.42e+04)}}= & 2.22e+04(1.22e+04) \\
F13 & 30& 4.14e+01(1.25e+01)+ & 6.14e+02(3.98e+02)+ & 2.04e+01(1.97e+00)+ & 9.75e+01(4.36e+01)+ & 1.03e+02(5.88e+01)+ & 4.36e+03(2.16e+03)+ & 1.94e+01(2.79e+00)+ & \cellcolor[rgb]{0.8,0.8,0.8}{\textbf{8.77e+00(2.39e+00)-}} & 1.27e+01(3.71e+00) \\
F14 & 30& 1.42e+02(1.85e-01)= & 1.41e+01(1.63e-01)= & 1.41e+01(2.63e-01)= & 1.42e+01(1.86e-01)= & 1.39e+01(2.43e-01)= & 1.42e+01(1.50e-01)= & 1.40e+01(2.92e-01)= & 1.39e+01(2.34e-01)= & \cellcolor[rgb]{0.8,0.8,0.8}{\textbf{1.39e+01(2.11e-01)}} \\
F15 & 30& 6.94e+02(7.41e+01)+ & 6.68e+02(8.04e+01)+ & \cellcolor[rgb]{0.8,0.8,0.8}{\textbf{4.03e+02(5.10e+01)}}- & 5.69e+02(8.90e+01)+ & 5.93e+02(9.82e+01)+ & 7.21e+02(1.10e+02)+ & 4.29e+02(5.58e+01)= & 5.16e+02(6.56e+01)+ & 4.67e+02(6.42e+01) \\

\hline

+/=/-&& 15/4/1& 18/2/0  & 14/4/2  & 17/3/0  &12/6/2 &16/2/2&6/11/3 & 6/11/3 & NA\\
Ranking & &6.90&7.90 & 4.85 &6.35& 4.25&6.40&3.50 &2.75& \cellcolor[rgb]{0.8,0.8,0.8}{\textbf{{2.10}}}\\

$p$-value   & & \cellcolor[rgb]{0.8,0.8,0.8}{\textbf{0.00}} & \cellcolor[rgb]{0.8,0.8,0.8}{\textbf{0.00}}& \cellcolor[rgb]{0.8,0.8,0.8}{\textbf{0.006}} & \cellcolor[rgb]{0.8,0.8,0.8}{\textbf{0.00}}& \cellcolor[rgb]{0.8,0.8,0.8}{\textbf{0.04}}& \cellcolor[rgb]{0.8,0.8,0.8}{\textbf{0.00}}&0.21&0.45
& NA\\

\hline
\hline
\end{tabular}}
\label{tab:30D_results}
\end{table*}
We compare the performance of \textbf{LLM-SAEA} with several state-of-the-art SAEAs, including ESAO (2019-TEVC)~\cite{wang2019novel}, IKAEA (2021-TEVC)~\cite{zhan2021fast}, TS-DDEO (2021-TCYB)~\cite{zhen2021two}, SA-MPSO~(2022-TSMCS)~\cite{liu2021surrogate}, SAMFEO (2023-TSMCS)~\cite{li2022expensive}, GL-SADE (2023-TCYB)~\cite{wang2022surrogate}, ESA (2023-TEVC)~\cite{zhen2022evolutionary}, and AutoSAEA (2024-TEVC)~\cite{xie2023surrogate}. For these comparison algorithms, we retain the parameter settings specified in their original publications. For our proposed \textbf{LLM-SAEA}, we set the population size $N$ to 100. Additionally, we utilize the GPT-3.5-turbo-0125 pre-trained LLM to act as both the decision-making and scoring experts. All experiments were performed on a computer with 32 GB of RAM and an Intel Core i7-10700 CPU running at 2.90 GHz.

For the \textbf{10D} benchmark problems presented in Table \ref{tab:10D_results}, \textbf{LLM-SAEA} surpasses all competing baseline algorithms on the majority of the test cases. Specifically, it significantly exceeds  ESAO, IKAEA, TS-DDEO, SA-MPSO, SAMFEO, GL-SADE, ESA, and AutoSAEA on 18, 17, 18, 16, 12, 17, 12, and 8 out of 20 benchmark problems, respectively. It only significantly underperforms IKAEA, SAMFEO, and AutoSAEA on 2, 3, and 4 problems, respectively. Moreover, \textbf{LLM-SAEA} achieves the best overall ranking among all the compared algorithms, according to the Friedman statistical test, demonstrating significantly superior performance compared to all other algorithms, except AutoSAEA.

For the \textbf{30D} test problems, \textbf{LLM-SAEA} demonstrates statistically significant superiority over ESAO, IKAEA, TS-DDEO, SA-MPSO, SAMFEO, GL-SADE, ESA, and AutoSAEA on 15, 18, 14, 17, 12, 16, 6, and 6 out of 20 benchmark problems, respectively. However, it is significantly surpassed by ESAO, IKAEA, TS-DDEO, SA-MPSO, SAMFEO, GL-SADE, ESA, and AutoSAEA on only 1, 0, 2, 0, 2, 2, 3, and 3 out of 20 problems, respectively. \textbf{LLM-SAEA} also achieves the highest overall ranking and shows significant superiority over all compared algorithms except ESA and AutoSAEA based on the Friedman test.

\begin{figure*}[t]
        \centering   
        \subfloat{\includegraphics[height=0.18in,width=6.8in]{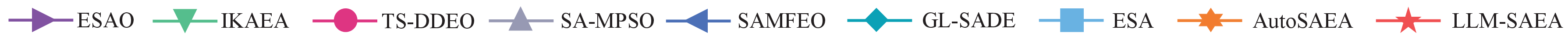}}\\
        \subfloat{\includegraphics[height=1.3in,width=1.7in]{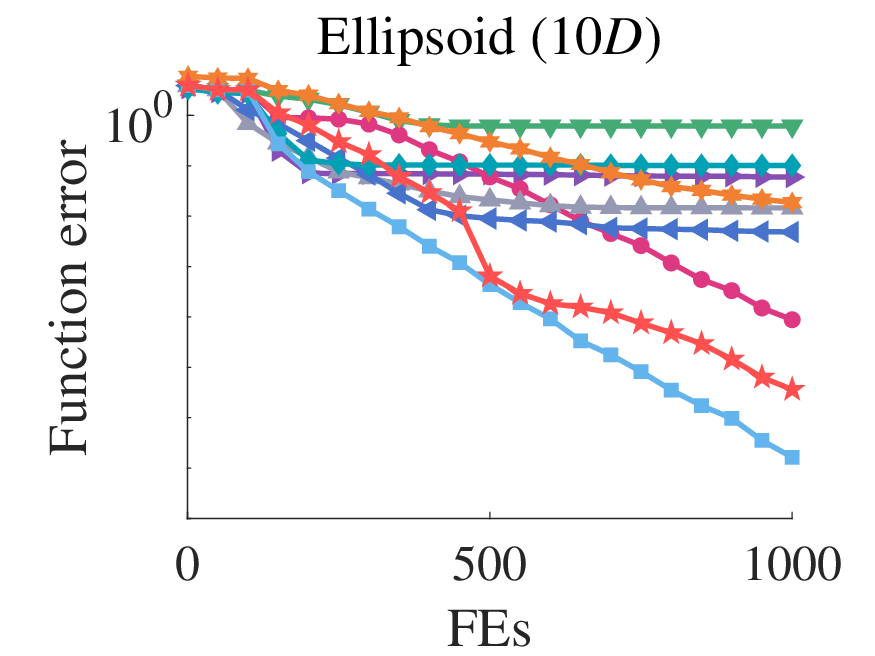}}
        \subfloat{\includegraphics[height=1.3in,width=1.7in]{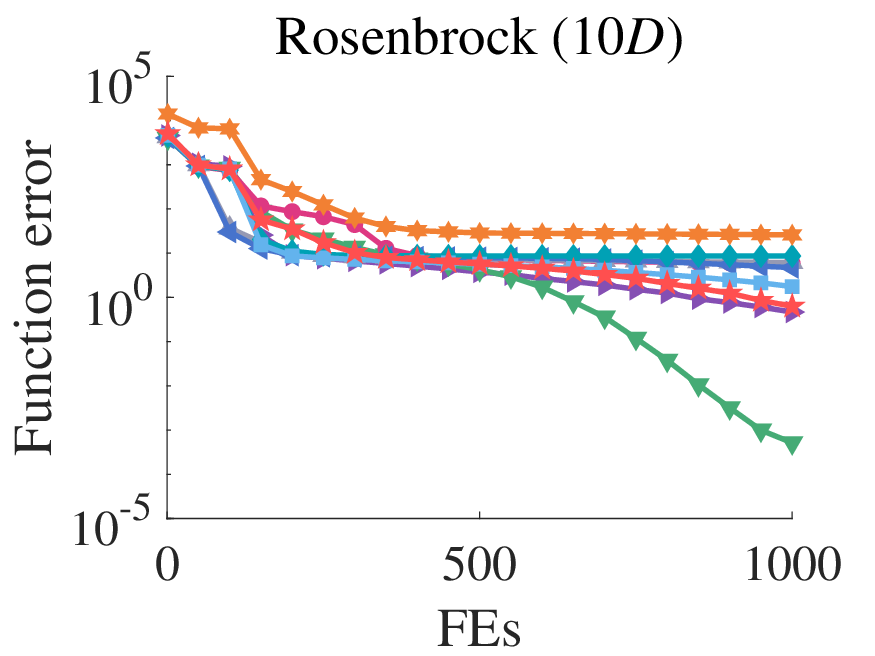}}
        \subfloat{\includegraphics[height=1.3in,width=1.7in]{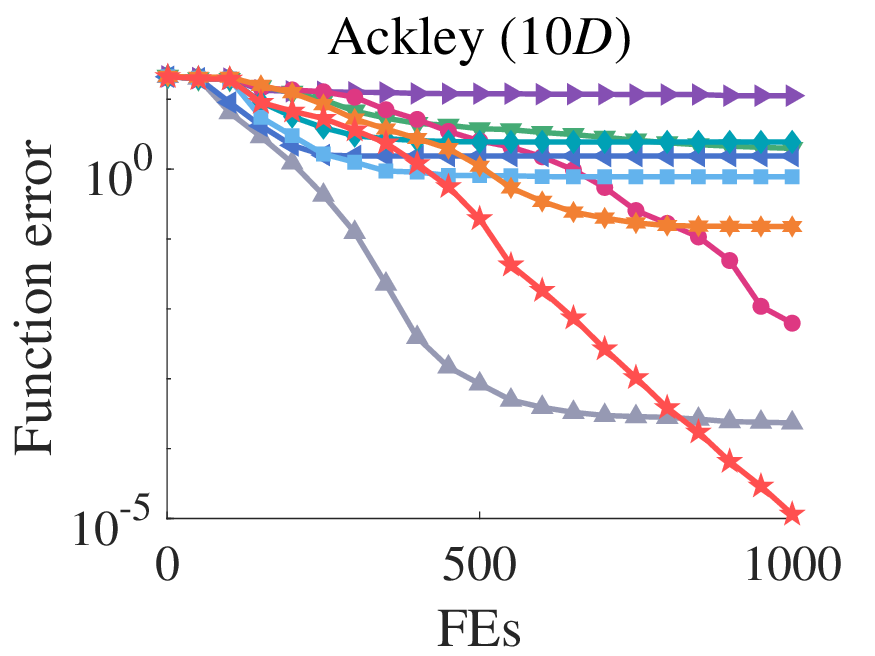}}
       \subfloat{\includegraphics[height=1.3in,width=1.7in]{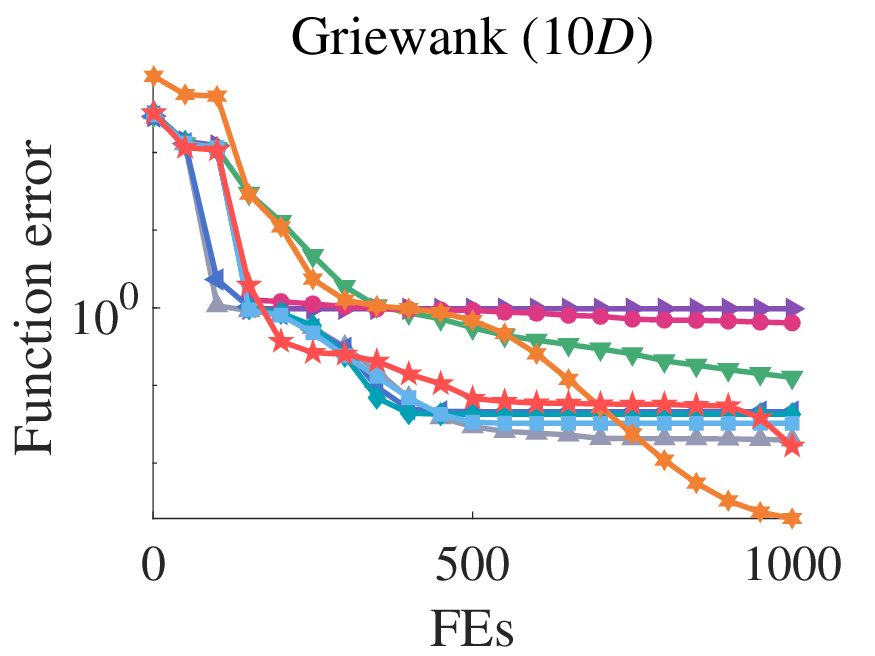}}\hfil
        \subfloat{\includegraphics[height=1.3in,width=1.7in]{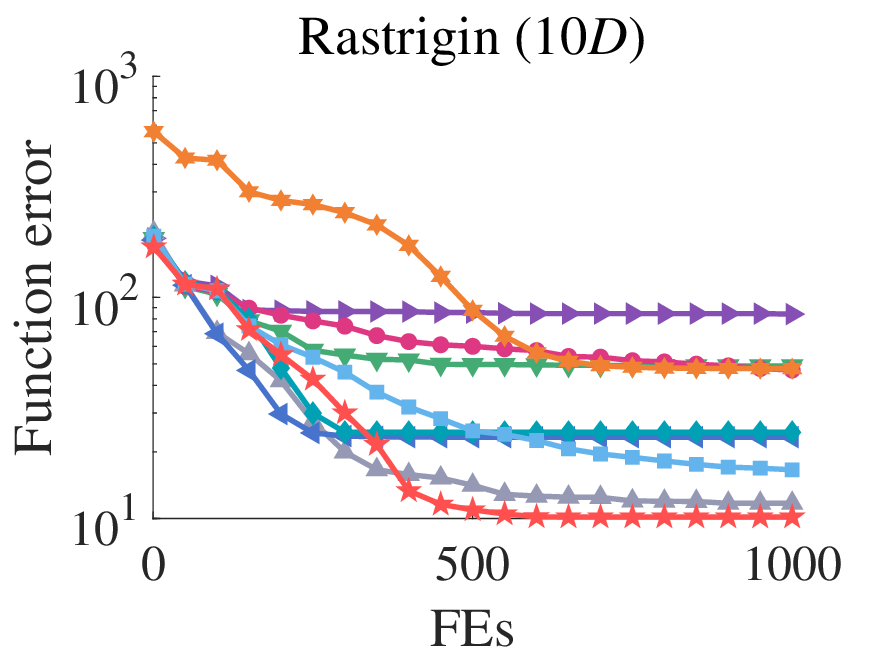}}
        \subfloat{\includegraphics[height=1.3in,width=1.7in]{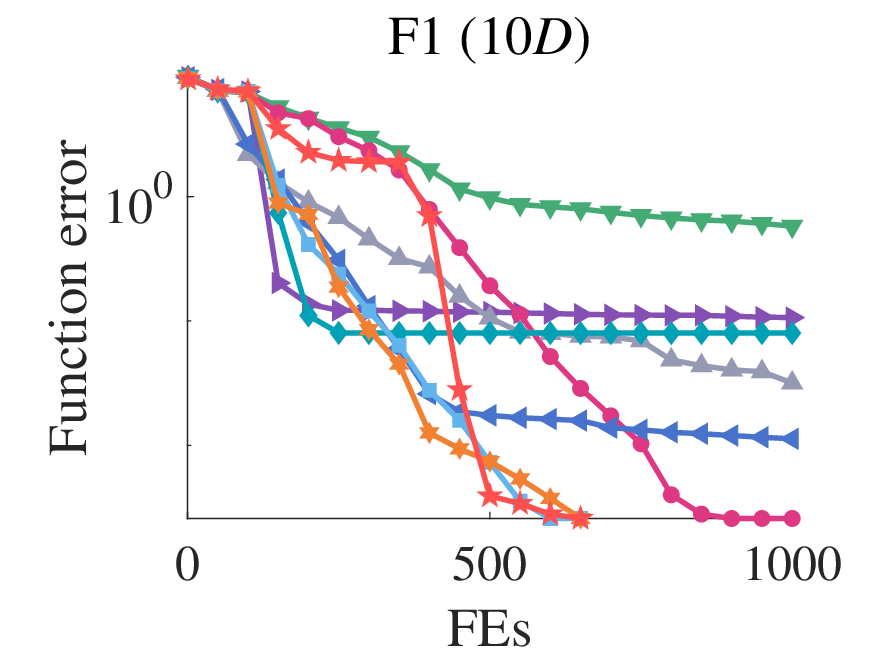}}
       \subfloat{\includegraphics[height=1.3in,width=1.7in]{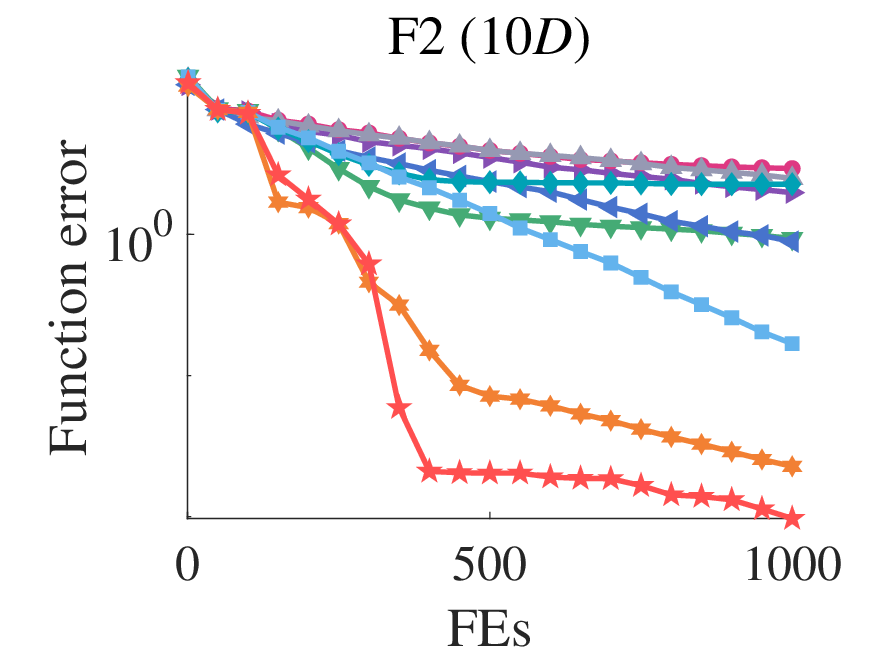}}
       \subfloat{\includegraphics[height=1.3in,width=1.7in]{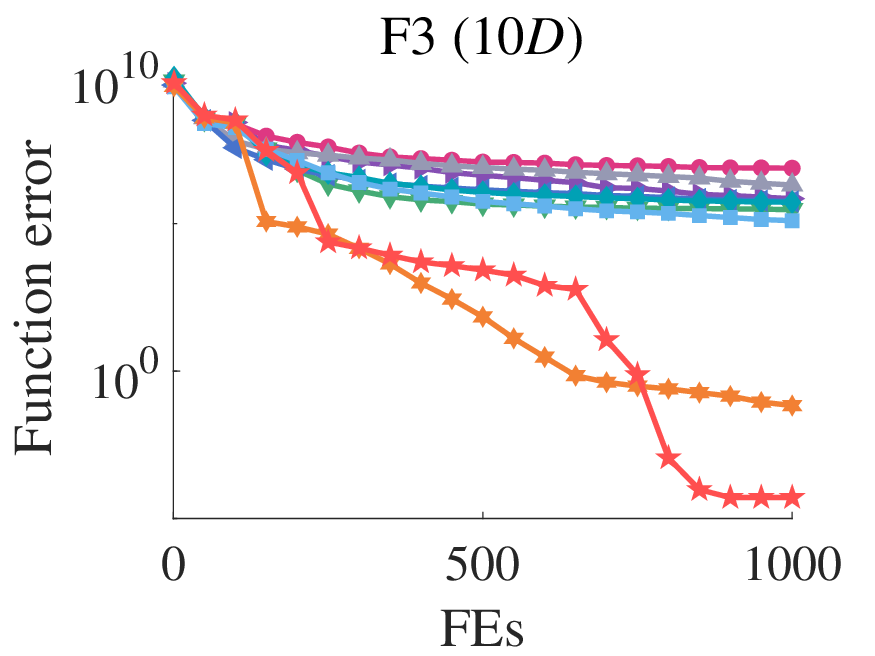}}
        \hfil
       \subfloat{\includegraphics[height=1.3in,width=1.7in]{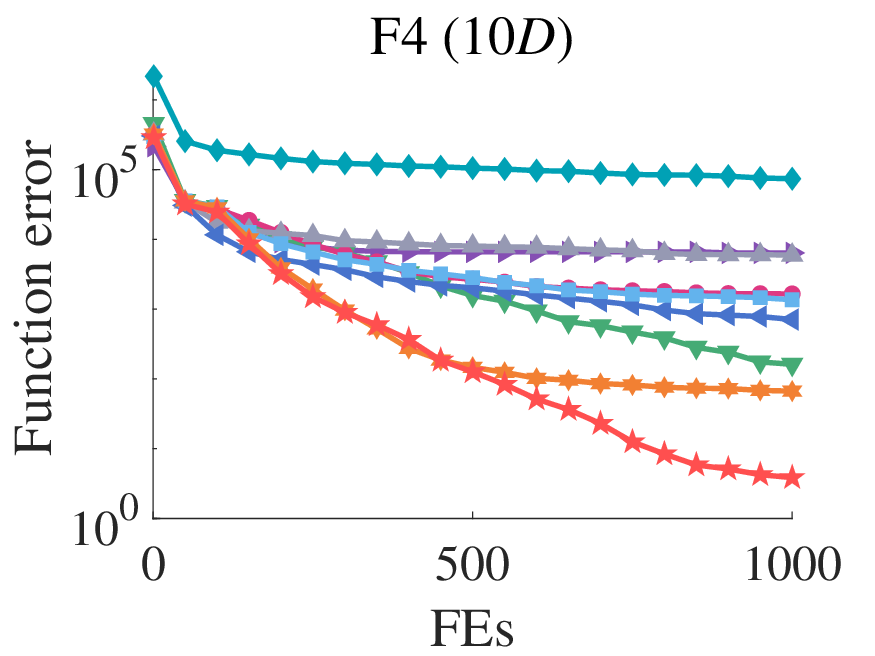}}
        \subfloat{\includegraphics[height=1.3in,width=1.7in]{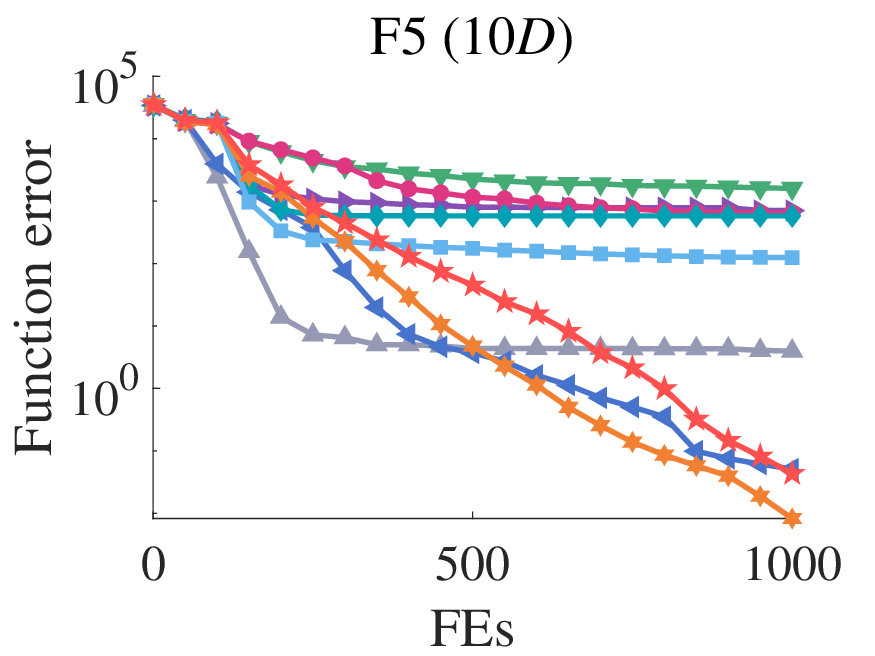}}
        \subfloat{\includegraphics[height=1.3in,width=1.7in]{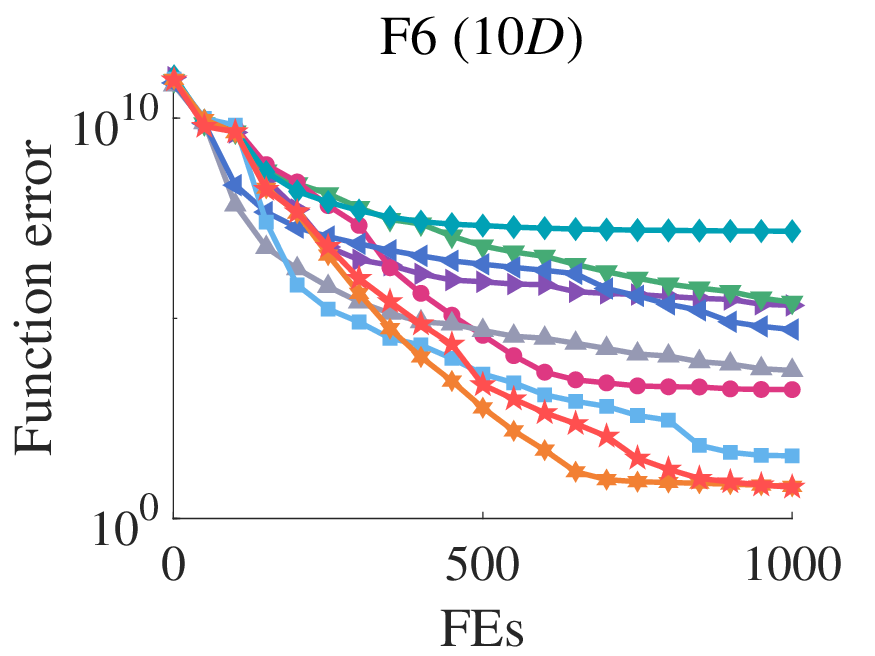}}
        \subfloat{\includegraphics[height=1.3in,width=1.7in]{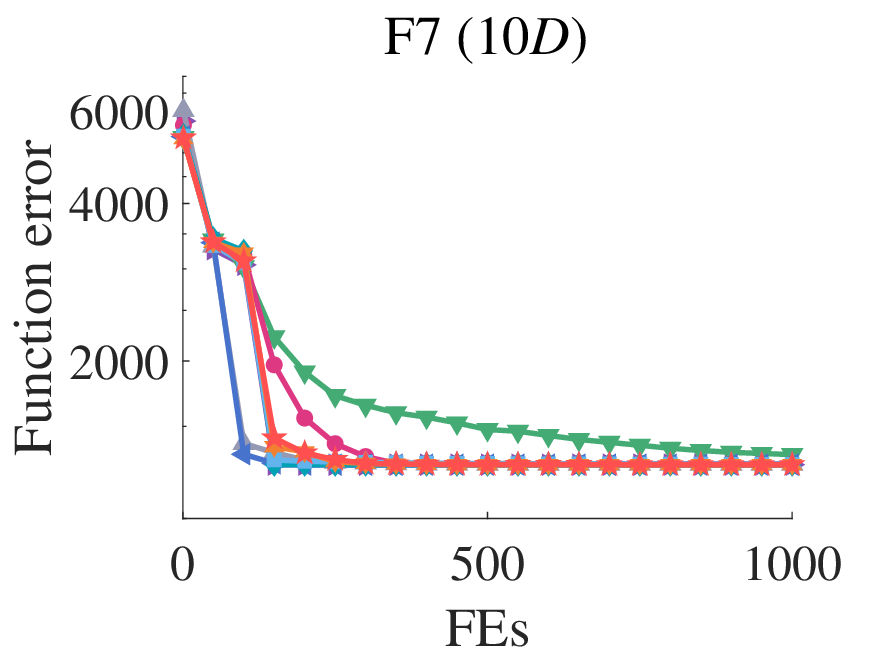}}
        \hfil
        \subfloat{\includegraphics[height=1.3in,width=1.7in]{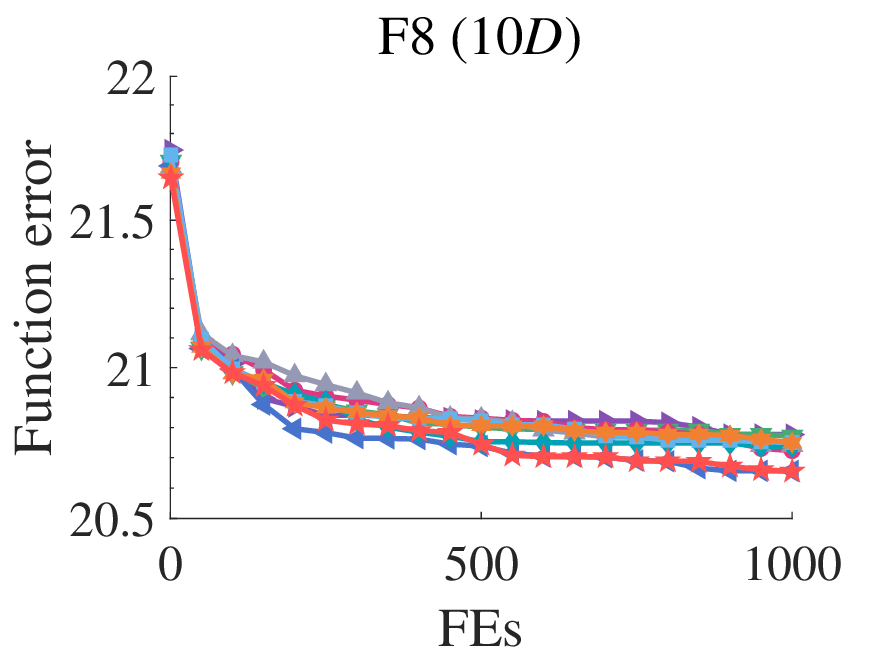}}
        \subfloat{\includegraphics[height=1.3in,width=1.7in]{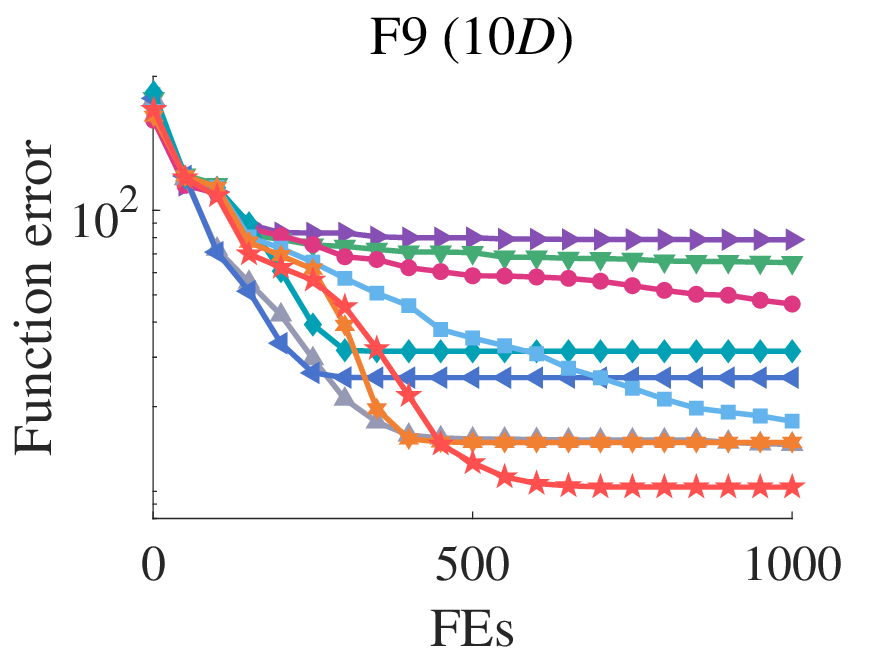}}
        \subfloat{\includegraphics[height=1.3in,width=1.7in]{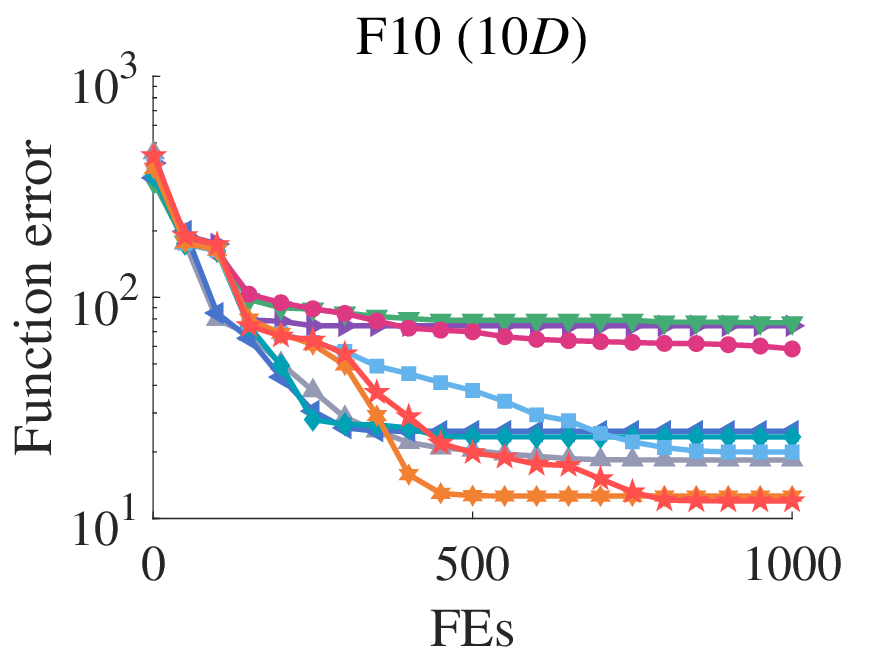}}
       \subfloat{\includegraphics[height=1.3in,width=1.7in]{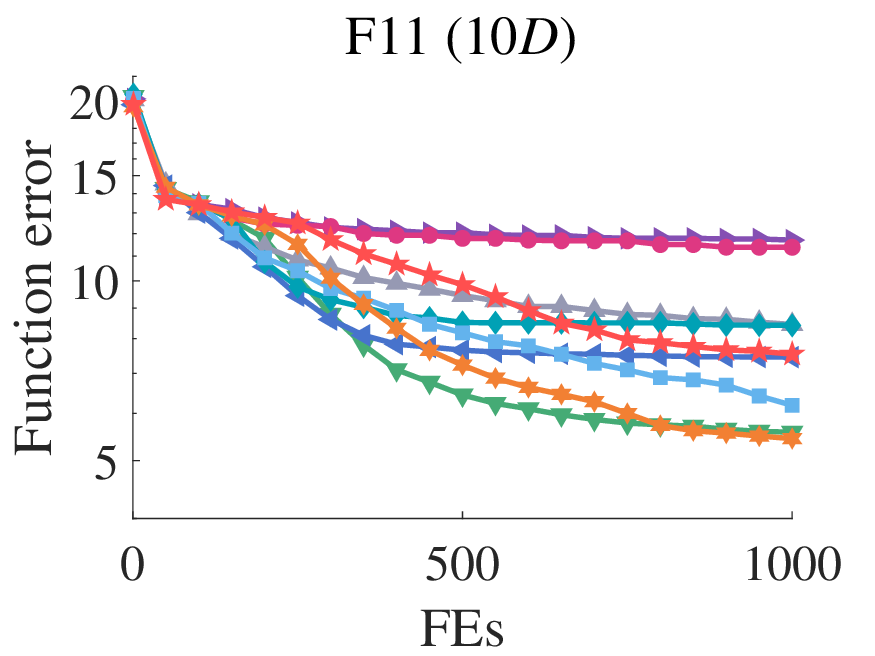}}
        \hfil
        \subfloat{\includegraphics[height=1.3in,width=1.7in]{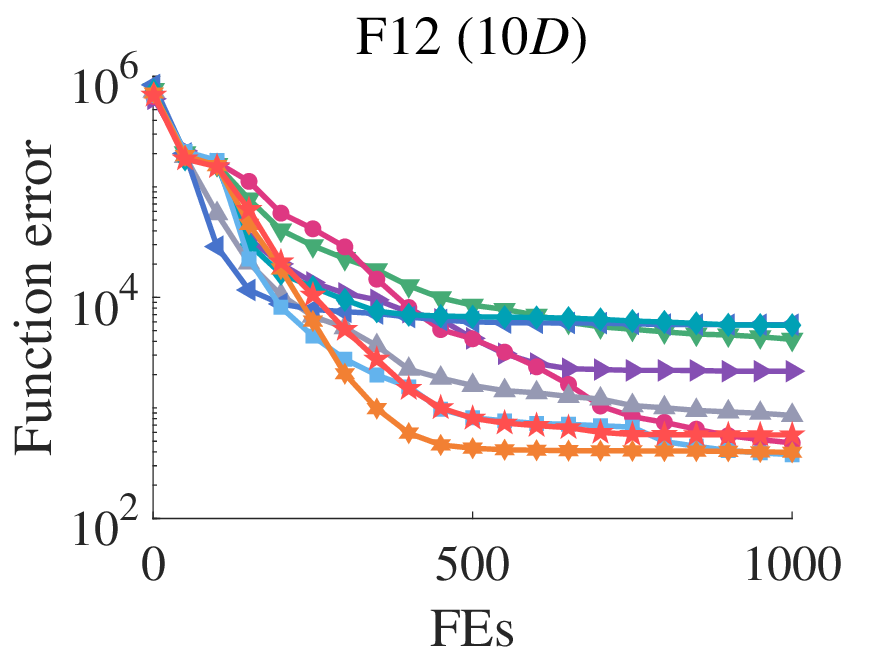}}
        \subfloat{\includegraphics[height=1.3in,width=1.7in]{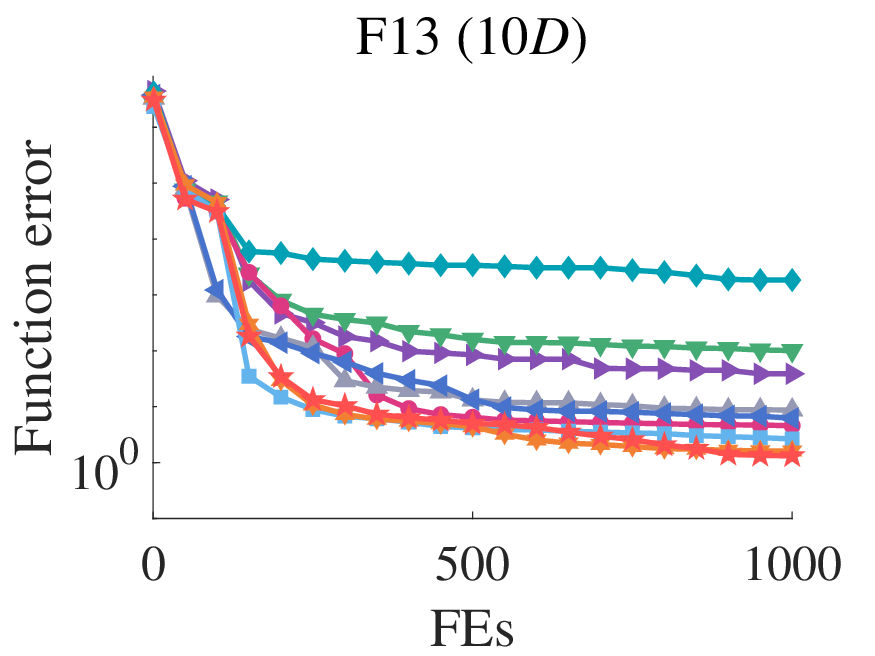}}
       \subfloat{\includegraphics[height=1.3in,width=1.7in]{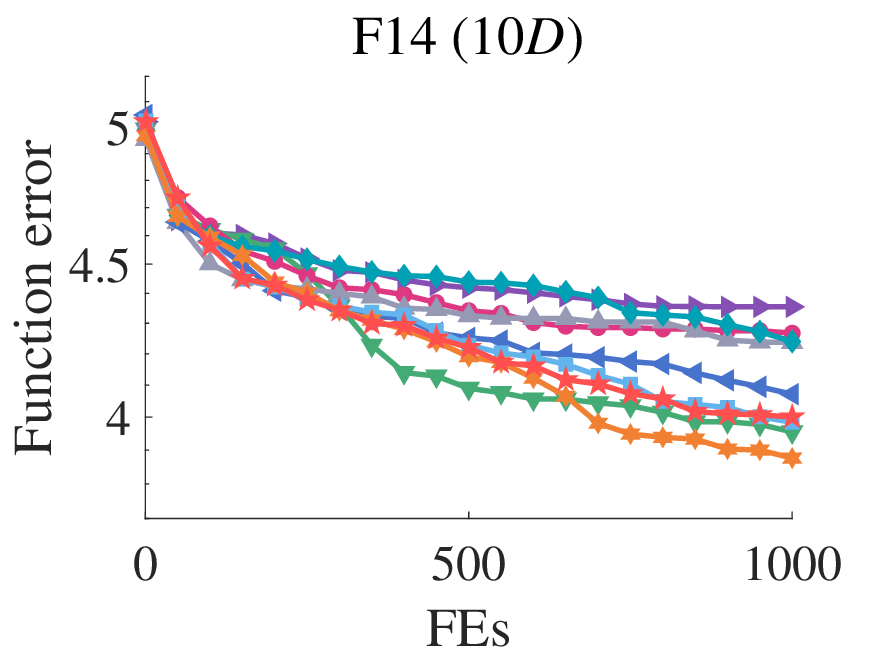}}
        \subfloat{\includegraphics[height=1.3in,width=1.7in]{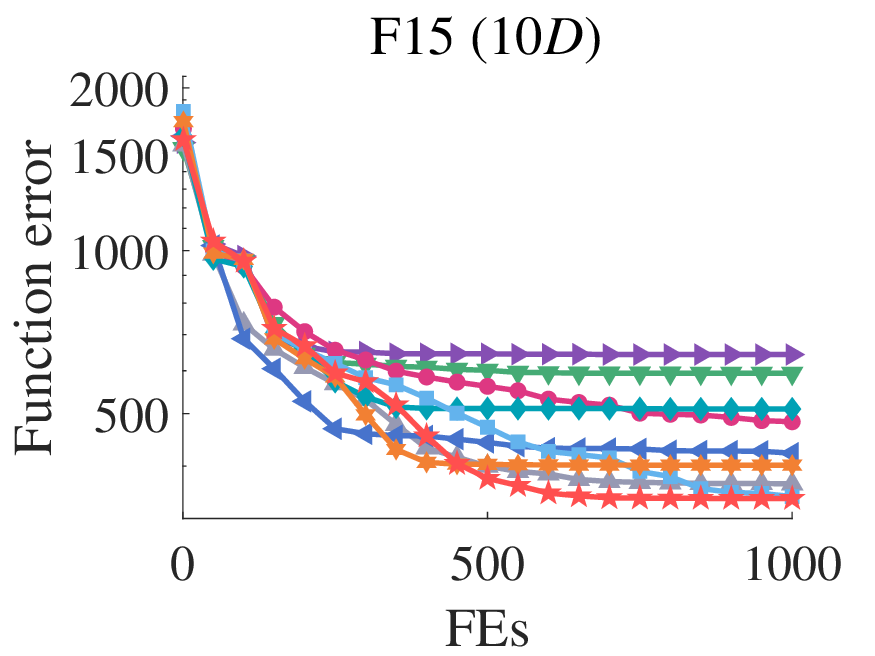}}

        \caption{The convergence curve of the mean function error value for each algorithm on \textbf{10D} problems.}       
        \label{fig: convergence}
\end{figure*}

Additionally, we plot the convergence curves of the average function error values for \textbf{LLM-SAEA} and its comparative algorithms on the \textbf{10D} test problems in Fig. \ref{fig: convergence}. The results clearly indicate that \textbf{LLM-SAEA} achieves superior performance across the majority of cases.

To elucidate the superior performance of \textbf{LLM-SAEA} over existing methods, we have the following remarks:
\begin{itemize}
\item Firstly, unlike static configuration SAEAs (e.g., ESAO, IKAEA, TS-DDEO, SA-MPSO, SAMFEO, and GL-SADE), \textbf{LLM-SAEA} can dynamically select the models and infill criteria based on the evolving contextual information during the optimization process. This mechanism significantly enhances adaptation and efficiency to various states, resulting in superior outcomes on most benchmark problems compared to these static configuration SAEAs.

\item Secondly, compared to dynamically configured SAEAs (e.g., ESA and AutoSAEA), \textbf{LLM-SAEA} manages dynamic configuration tasks by conveying them to LLMs via natural language. This approach enhances configuration efficiency and mitigates potential performance degradation caused by human-designed algorithmic component preferences. However, it should be noted that the performance of \textbf{LLM-SAEA} does not show a statistically significant improvement over ESA and AutoSAEA according to the Friedman test. Therefore, further leveraging LLMs to enhance the performance of SAEAs remains a key focus of our future research.
\end{itemize}

\subsection{Effectiveness Analysis of LLM-SAEA}
\subsubsection{Effectiveness Analysis of LLM-SAEA Dynamic Configuration Capabilty}

\begin{table*}[!h]
\renewcommand{\arraystretch}{1.1}
\tiny
\caption{The average and standard deviation of the function error values for \textbf{LLMSAEA} and its variants on the F1-F15 (\textbf{10D}) benchmark problems.
}
\centering
\resizebox{\textwidth}{!}{\begin{tabular}{p{0.7cm}p{0.1cm}p{1.90cm}p{1.90cm}p{1.90cm}p{1.90cm}p{1.85cm}p{1.85cm}p{1.85cm}p{1.85cm}p{1.85cm}p{1.85cm}p{1.85cm}p{1.85cm}p{1.85cm}}
\hline
\hline
Problem & D& V-A1& V-A2 & V-A3 & V-A4 & V-A5 & V-A6 & V-A7 & V-A8 & V-Seq & V-Random & V-Alter & V-Q & LLM-SAEA \\

F1 &10& 3.03e-14(6.03e-14)+ & 3.79e-15(1.47e-14)= & 7.69e-13(4.27e-13)+ & 2.37e-04(6.11e-04)+ & 4.43e-10(8.03e-10)+ & 8.41e-13(3.14e-13)= & 3.02e+02(2.97e+02)+ & 3.78e+03(1.61e+03)+ & 8.26e-13(8.44e-13)+ & \cellcolor[rgb]{0.8,0.8,0.8}{\textbf{0.00e+00(0.00e+00)}}= & \cellcolor[rgb]{0.8,0.8,0.8}{\textbf{0.00e+00(0.00e+00)}}= & \cellcolor[rgb]{0.8,0.8,0.8}{\textbf{0.00e+00(0.00e+00)}}= & \cellcolor[rgb]{0.8,0.8,0.8}{\textbf{0.00e+00(0.00e+00)}} \\

F2 &10& 3.55e-02(7.07e-02)+ & 2.50e-01(1.63e-01)+ & 5.24e-01(8.00e-01)+ & 2.95e+03(2.77e+03)+ & 1.69e-07(8.22e-08)+ & \cellcolor[rgb]{0.8,0.8,0.8}{\textbf{7.09e-12(3.04e-12)}}= & 5.73e+03(2.19e+03)+ & 1.17e+04(4.24e+03)+ & 5.61e+02(2.67e+02)+ & 1.67e-11(1.59e-11)- & 1.90e-11(3.20e-11)- & 9.60e-12(1.14e-11)- & 8.57e-11(2.09e-10) \\

F3 &10& 2.77e+05(2.82e+05)+ & 2.83e+05(2.46e+05)+ & 7.53e+05(8.25e+05)+ & 3.79e+07(2.32e+07)+ & 1.34e+02(4.41e+02)+ & \cellcolor[rgb]{0.8,0.8,0.8}{\textbf{5.06e-07(6.96e-07)}}- & 2.20e+07(1.76e+07)+ & 7.86e+07(3.94e+07)+ & 4.57e+06(1.97e+06)+ & 6.14e-06(4.54e-06)= & 1.57e-05(2.47e-05)= & 9.71e-06(9.44e-06)= & 5.13e-05(9.50e-05) \\

F4 &10& 7.77e+02(1.03e+03)+ & 2.71e+02(4.52e+02)+ & 4.28e+01(1.53e+02)+ & 3.51e+03(5.64e+03)+ &\cellcolor[rgb]{0.8,0.8,0.8}{\textbf{1.18e-02(4.48e-02)}}- & 1.42e+03(1.46e+03)+ & 5.45e+03(1.77e+03)+ & 1.12e+04(2.88e+03)+ & 9.14e+02(7.12e+02)+ & 6.87e-01(1.00e+00)+ & 2.04e-01(1.28e-01)+ & 3.15e-01(2.10e-01)+ & 3.85e+00(1.19e+01) \\

F5 &10& 1.98e-01(4.89e-01)+ & 4.46e-02(2.75e-02)= & 1.02e+02(3.96e+02)= & 8.02e+02(4.54e+02)+ & 2.16e+03(7.12e+02)+ & 3.76e+03(2.80e+03)+ & 8.33e+03(1.98e+03)+ & 8.28e+03(1.67e+03)+ & 5.42e-01(3.72e-01)+ & 7.67e-02(5.46e-02)+ & 4.84e-02(3.10e-02)+ & 8.42e-02(5.37e-02)+ & \cellcolor[rgb]{0.8,0.8,0.8}{\textbf{4.32e-02(7.81e-02)}}\\

F6 &10& 1.55e+01(5.16e+01)- & 4.73e+00(1.55e+00)- & 
\cellcolor[rgb]{0.8,0.8,0.8}{\textbf{4.63e+00(1.81e+00)}}- & 2.08e+02(7.39e+02)+ & 4.81e+01(4.74e+01)+ & 
4.62e+07(5.99e+07)+ & 5.84e+06(1.03e+07)+ & 2.39e+08(1.61e+08)+ & 9.27e+02(1.04e+03)+ & 8.54e+00(4.54e+00)+ & 8.78e+00(3.68e+00)+ & 2.06e+01(4.41e+01)+ & 6.04e+00(1.18e+00) \\

F7 &10& 1.30e+03(9.76e+00)+ & 1.30e+03(1.64e+01)+ & 1.27e+03(1.12e+00)+ & 1.29e+03(2.24e+00)+ & 1.27e+03(3.79e+00)+ & 1.29e+03(2.09e+00)+ & 2.06e+03(1.44e+02)+ & 1.68e+03(9.94e+01)+ & 1.27e+03(1.63e-01)+ & 1.27e+03(2.60e-01)= & 1.27e+03(4.87e-01)= & 1.27e+03(8.69e-02)= & \cellcolor[rgb]{0.8,0.8,0.8}{\textbf{1.27e+03(2.00e-01)}} \\

F8 &10& 2.07e+01(1.17e-01)= & 2.07e+01(1.11e-01)= & 2.07e+01(1.17e-01)= & 2.08e+01(9.04e-02)+ & 2.08e+01(9.95e-02)+ & 2.08e+01(1.13e-01)= & 2.07e+01(1.16e-01)= & 2.07e+01(1.34e-01)= & 2.07e+01(1.17e-01)= & 2.08e+01(1.24e-01)+ & 2.07e+01(1.19e-01)= & 2.09e+01(9.40e-02)+ & \cellcolor[rgb]{0.8,0.8,0.8}{\textbf{2.07e+01(1.03e-01)}} \\
F9 &10&1.74e+01(9.21e+00)+ & 1.92e+01(6.16e+00)+ & 4.12e+01(1.27e+01)+ & 6.05e+01(1.00e+01)+ & 5.46e+01(1.43e+01)+ & 7.37e+01(1.33e+01)+ & 1.90e+01(7.85e+00)+ & 7.67e+01(1.50e+01)+ & 1.69e+01(6.60e+00)+ & 1.09e+01(2.96e+00)= & 1.13e+01(4.02e+00)= & 1.04e+01(3.53e+00)= & \cellcolor[rgb]{0.8,0.8,0.8}{\textbf{1.03e+01(2.98e+00)}} \\

F10 &10& 2.59e+01(1.30e+01)+ & 2.22e+01(7.21e+00)+ & 4.75e+01(8.14e+00)+ & 5.86e+01(8.43e+00)+ & 5.43e+01(1.05e+01)+ & 7.16e+01(8.35e+00)+ & 2.45e+01(7.90e+00)+ & 1.00e+02(1.42e+01)+ & 1.56e+01(6.50e+00)= & 1.51e+01(9.14e+00)= &
\cellcolor[rgb]{0.8,0.8,0.8}{\textbf{9.68e+00(4.62e+00)}}= & 1.35e+01(6.21e+00)= & 1.20e+01(6.14e+00) \\

F11 &10& \cellcolor[rgb]{0.8,0.8,0.8}{\textbf{6.27e+00(1.25e+00)}}= & 6.78e+00(1.35e+00)= & 1.08e+01(2.31e+00)+ & 1.16e+01(1.07e+00)+ & 1.17e+01(1.07e+00)+ & 1.27e+01(9.56e-01)+ & 8.01e+00(1.65e+00)= & 1.15e+01(1.14e+00)+ & 9.41e+00(1.85e+00)+ & 7.00e+00(2.38e+00)= & 7.83e+00(1.94e+00)= & 6.52e+00(2.58e+00)= & 7.54e+00(2.24e+00) \\

F12 &10& 3.24e+02(5.58e+02)= & 4.71e+02(7.02e+02)= & 3.93e+02(5.07e+02)= & 3.84e+04(1.72e+04)+ & 1.44e+04(8.40e+03)+ & 8.94e+04(2.92e+04)+ & 1.21e+04(9.68e+03)+ & 6.61e+04(2.13e+04)+ & 2.48e+03(3.80e+03)+ & 6.80e+02(4.89e+02)= & 
\cellcolor[rgb]{0.8,0.8,0.8}{\textbf{3.06e+02(5.64e+02)}}= & 3.82e+02(6.04e+02)= & 5.70e+02(7.16e+02) \\

F13 &10& 1.68e+00(7.62e-01)+ & 1.42e+00(6.56e-01)= & 3.79e+00(8.47e-01)+ & 1.02e+01(4.59e+00)+ & 4.88e+00(8.85e-01)+ & 3.54e+01(2.71e+01)+ & 2.23e+00(1.21e+00)+ & 6.09e+03(8.11e+03)+ & 6.83e+00(1.59e+00)+ & 2.69e+00(1.20e+00)+ & 2.59e+00(1.41e+00)+ & 2.81e+00(1.46e+00)+ &
\cellcolor[rgb]{0.8,0.8,0.8}{\textbf{1.32e+00(1.44e+00)}}\\

F14 &10& 4.13e+00(3.68e-01)= & 4.05e+00(2.58e-01)= & 4.12e+00(1.71e-01)= & 4.35e+00(1.12e-01)+ & 4.15e+00(2.90e-01)+ & 4.41e+00(2.02e-01)+ & 
\cellcolor[rgb]{0.8,0.8,0.8}{\textbf{3.85e+00(1.68e-01)}}- & 4.36e+00(1.39e-01)+ & 4.18e+00(2.18e-01)+ & 4.12e+00(2.19e-01)= & 4.01e+00(1.58e-01)= & 4.04e+00(2.06e-01)= & 4.00e+00(1.99e-01) \\

F15 &10& 3.59e+02(1.20e+02)= & 3.54e+02(1.36e+02)= & 4.20e+02(1.52e+02)+ & 5.02e+02(6.62e+01)+ & 5.68e+02(1.55e+02)+ & 7.79e+02(7.46e+01)+ & 5.36e+02(1.41e+02)+ & 7.58e+02(7.80e+01)+ & 5.45e+02(4.25e+02)+ & 3.52e+02(1.55e+02)= & 
\cellcolor[rgb]{0.8,0.8,0.8}{\textbf{3.19e+02(1.13e+02)}}= & 3.75e+02(1.21e+02)= & 3.48e+02(1.21e+02) \\
\hline

+/=/-&& 9/5/1& 6/8/1  & 10/4/1  & 15/0/0  &14/0/1 &11/3/1&12/2/1 & 14/1/0 & 13/2/0& 5/9/1& 4/10/1& 5/9/1& NA\\

Ranking && 5.67 & 5.33 & 7.07 & 10.73 & 8.40 & 10.00 & 9.13 & 11.87 & 7.73 & 4.60 & 3.33 & 3.73 & \cellcolor[rgb]{0.8,0.8,0.8}{\textbf{2.60}} \\

$p$-value && 0.22 & 0.30 & \cellcolor[rgb]{0.8,0.8,0.8}{\textbf{0.01}} & \cellcolor[rgb]{0.8,0.8,0.8}{\textbf{0.00}} & \cellcolor[rgb]{0.8,0.8,0.8}{\textbf{0.0004}} & \cellcolor[rgb]{0.8,0.8,0.8}{\textbf{0.00}} & \cellcolor[rgb]{0.8,0.8,0.8}{\textbf{0.0001}} & \cellcolor[rgb]{0.8,0.8,0.8}{\textbf{0.00}} & \cellcolor[rgb]{0.8,0.8,0.8}{\textbf{0.003}} & 0.37 & 0.82 & 0.82 & NA \\
\hline
\hline
\end{tabular}}
\label{tab:dynamic_configuration_10D_results}
\end{table*}

 \begin{figure*}[!h]
        \centering
\includegraphics[height=1.7 in,width=6.65in]{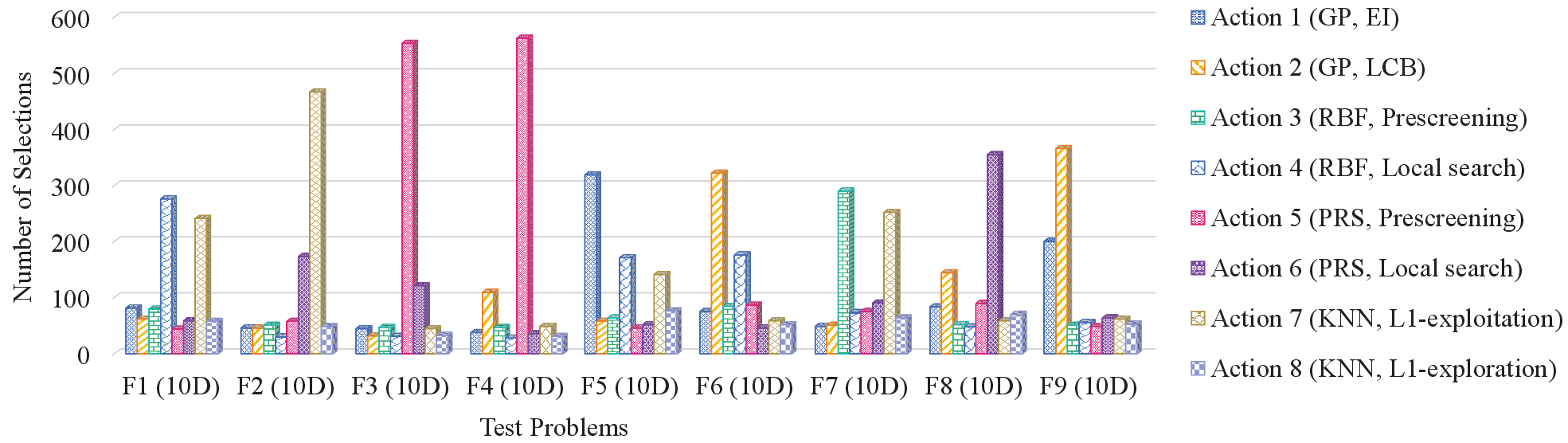}
        \caption{The frequency of each action being selected throughout the optimization process of \textbf{LLM-SAEA} on the F1-F9 (\textbf{D=10}) problems.}
        \label{fig: Times of actions}
\end{figure*}

\textbf{LLM-SAEA} employs a collaboration-of-experts framework, with the LLM serving as both the scoring expert and the decision expert to dynamically configure the surrogate model and infill sampling criteria. To further assess  its effectiveness, we compared \textbf{LLM-SAEA} against the following 12 variants:

\begin{itemize}
\item V-A$i$ $(i=1,\ldots,8)$: \textbf{LLM-SAEA} employs only the $i$-th action in the combinatorial action set $\mathcal{CA}$ throughout the entire optimization process.

\item V-Seq: \textbf{LLM-SAEA} sequentially selects an action from the combinatorial action set $\mathcal{CA}$ during the optimization process.

\item V-Random: \textbf{LLM-SAEA} randomly selects an action from the combinatorial action set $\mathcal{CA}$ in each generation.

\item V-Alter: \textbf{LLM-SAEA} alternately selects an action from the combinatorial action set $\mathcal{CA}$~\cite{wang2019novel}. Specifically, if the selected action provides a better solution than the current best, it continues with this action in the next generation. Otherwise, it randomly selects a different action from set $\mathcal{CA}$.

\item V-Q: \textbf{LLM-SAEA} employs Q-learning, developed in ESA~\cite{zhen2022evolutionary}, to select an action from the combinatorial action set $\mathcal{CA}$ in each iteration.

\end{itemize}

The detailed results of \textbf{LLM-SAEA} and the 12 compared variants on the F1-F15 (\textbf{10D}) benchmark problems are provided in Table \ref{tab:dynamic_configuration_10D_results}.  As shown in Table \ref{tab:dynamic_configuration_10D_results}, \textbf{LLM-SAEA} significantly outperforms V-A1, V-A2, V-A3, V-A4, V-A5, V-A6, V-A7, V-A8, V-Seq, V-Rand, V-Alter, and V-Q on 9, 6, 10, 15, 14, 11, 12, 14, 13, 5, 4, and 5 out of 15 problems, respectively, and is only inferior on 1, 1, 1, 0, 1, 1, 1, 0, 0, 1, 1, and 1 problem, respectively. It achieved the best average ranking, and according to the Friedman test, it also significantly outperforms the V-A3, V-A4, V-A5, V-A6, V-A7, V-A8, and V-Seq variants. These results confirm the effectiveness of LLMs in the dynamic configuration of SAEAs. Additionally, we collect data about the number of times each action is selected throughout the optimization process of \textbf{LLM-SAEA} on the F1-F9 (\textbf{10D}) test problems, as shown in Fig. \ref{fig: Times of actions}. Visually, across different optimization problems, \textbf{LLM-SAEA} demonstrates the capability to dynamically select appropriate models and infill criteria, further validating its adaptive performance.

\subsubsection{Effectiveness of Collaboration-of-Experts Framework}

\begin{table}[!h]
\renewcommand{\arraystretch}{1.1}
\tiny
\caption{The average function error values and their standard deviations for all compared algorithms on the F1-F15 (\textbf{10D}) benchmark problems.}
\centering
{\begin{tabular}
{p{0.7cm}p{0.1cm}p{1.85cm}p{1.85cm}p{1.85cm}}

\hline
\hline
Problem & D  & V-DE & LLM-SAEA \\

\hline
F1 & 10 & 1.71e-13(4.71e-13) + & \cellcolor[rgb]{0.8,0.8,0.8}{\textbf{0.00e+00(0.00e+00)}}  \\

F2 & 10 & 1.97e-08(1.39e-08) + & \cellcolor[rgb]{0.8,0.8,0.8}{\textbf{8.57e-11(2.09e-10)}}  \\

F3 & 10 & 9.33e+01(1.47e+02)+ & \cellcolor[rgb]{0.8,0.8,0.8}{\textbf{5.13e-05(9.50e-05)}}  \\

F4 & 10 & 4.18e+02(1.25e+03)= & \cellcolor[rgb]{0.8,0.8,0.8}{\textbf{3.85e+00(1.19e+01)}}  \\

F5 & 10 & 1.10e-01(1.63e-01)= & \cellcolor[rgb]{0.8,0.8,0.8}{\textbf{4.32e-02(7.81e-02)}}  \\

F6 & 10 & 7.91e+00(2.27e+00)+ & \cellcolor[rgb]{0.8,0.8,0.8}{\textbf{6.04e+00(1.18e+00)}}  \\

F7 & 10 & \cellcolor[rgb]{0.8,0.8,0.8}{\textbf{1.27e+03(4.23e-02)}}= & 1.27e+03(2.00e-01)  \\

F8 & 10 & 2.07e+01(1.69e-01)= & \cellcolor[rgb]{0.8,0.8,0.8}{\textbf{2.07e+01(1.03e-01)}}  \\

F9 & 10 &1.15e+01(4.92e+00)= & \cellcolor[rgb]{0.8,0.8,0.8}{\textbf{1.03e+01(2.98e+00)}}  \\

F10 & 10 & 1.55e+01(1.17e+01)= & \cellcolor[rgb]{0.8,0.8,0.8}{\textbf{1.20e+01(6.14e+00)}}  \\

F11 & 10 &7.61e+00(1.98e+00)= & \cellcolor[rgb]{0.8,0.8,0.8}{\textbf{7.54e+00(2.24e+00)}}  \\

F12 & 10 &\cellcolor[rgb]{0.8,0.8,0.8}{\textbf{2.17e+02(4.99e+02)}}= & 5.70e+02(7.16e+02)  \\

F13 & 10 &1.42e+00(6.78e-01)= & \cellcolor[rgb]{0.8,0.8,0.8}{\textbf{1.32e+00(1.44e+00)}}  \\

F14 & 10 &4.08e+00(2.62e-01)= & \cellcolor[rgb]{0.8,0.8,0.8}{\textbf{4.00e+00(1.99e-01)}}  \\

F15 & 10 &4.33e+02(7.71e+01)= & \cellcolor[rgb]{0.8,0.8,0.8}{\textbf{3.48e+02(1.21e+02)}}  \\
\hline
+/=/- & & 4/11/0 & NA \\

Ranking &  &1.8 & \cellcolor[rgb]{0.8,0.8,0.8}{\textbf{1.2}} \\

$p$-value & &  \cellcolor[rgb]{0.8,0.8,0.8}{\textbf{0.02}} & NA \\
\hline
\hline
\end{tabular}}
\label{tab:Statistics_Effextness_framework_10D_results}
\end{table}

In this study, we decompose the dynamic algorithm configuration of SAEAsinto two sub-tasks: decision-making (\textbf{LLM-DE}) and scoring (\textbf{LLM-SE}). Furthermore, we compare our approach, which utilizes both LLMs, with a variant that employs only \textbf{LLM-DE}. The detailed results are presented in Table \ref{tab:Statistics_Effextness_framework_10D_results}. As shown in Table \ref{tab:Statistics_Effextness_framework_10D_results}, \textbf{LLM-SAEA} significantly outperforms the variant on 4 out of 15 problems, 
with no significant underperformance on any problem. Additionally, according to the Friedman test, \textbf{LLM-SAEA} demonstrates superior performance compared to the \textbf{LLM-DE} variant. Overall, these results confirm the effectiveness of using both LLMs within our framework.

\subsubsection{Effectiveness of Self-reflection Component}
A self-reflection component (SRC) is embedded in \textbf{LLM-DE} to perform self-reflection on its selected results. To assess its effectiveness, we benchmark \textbf{LLM-SAEA} against the two variants listed below:

\begin{itemize}
\item V-WoSRC:  In this variant, \textbf{LLM-DE} outputs only a final set of recommended actions $\mathcal{A}^*$ without an accompanying set of confidence labels by the SRC.

\item V-SRC-Certain: The recommended action set $\mathcal{A}^*$ in each iteration includes only actions labeled as 'certain' by the SRC of \textbf{LLM-DE}.
\end{itemize}






\begin{table}[!h]
\renewcommand{\arraystretch}{1.1}
\tiny
\caption{The average and standard deviation of the function error across all algorithms compared on the F1-F15 (\textbf{10D}) benchmark problems.
}
\centering
{\begin{tabular}
{p{0.7cm}p{0.1cm}p{1.85cm}p{1.85cm}p{1.85cm}}

\hline
\hline
Problem & D & V-WoSRC & V-SRC-Certain & LLM-SAEA \\

\hline
F1 & 10 & 1.75e-02(4.95e-02)+ & 2.86e+02(7.15e+02)+ & \cellcolor[rgb]{0.8,0.8,0.8}{\textbf{0.00e+00(0.00e+00)}}  \\

F2 & 10 & 2.12e-01(5.98e-01)+ & 2.12e+03(5.99e+03)+ & \cellcolor[rgb]{0.8,0.8,0.8}{\textbf{8.57e-11(2.09e-10)}}  \\

F3 & 10 & 1.15e+06(2.57e+06)+ & 9.29e+06(1.83e+07)= & \cellcolor[rgb]{0.8,0.8,0.8}{\textbf{5.13e-05(9.50e-05)}}  \\

F4 & 10 & \cellcolor[rgb]{0.8,0.8,0.8}{\textbf{3.70e+00(8.19e+00)}}= & 3.50e+03(3.56e+03)+ & 3.85e+00(1.19e+01)  \\

F5 & 10 & 4.22e+02(7.21e+02)+ & 1.93e+03(3.57e+03)+ & \cellcolor[rgb]{0.8,0.8,0.8}{\textbf{4.32e-02(7.81e-02)}}  \\

F6 & 10 & 1.61e+03(4.45e+03)= & 1.06e+08(1.86e+08)= & \cellcolor[rgb]{0.8,0.8,0.8}{\textbf{6.04e+00(1.18e+00)}}  \\

F7 & 10 & 1.27e+03(4.28e+00)+ & 1.27e+03(5.57e+00)+ & \cellcolor[rgb]{0.8,0.8,0.8}{\textbf{1.27e+03(2.00e-01)}}  \\

F8 & 10 & 2.07e+01(9.81e-02)= & 2.07e+01(1.11e-01)= & \cellcolor[rgb]{0.8,0.8,0.8}{\textbf{2.07e+01(1.03e-01)}}  \\

F9 & 10 & 1.18e+01(8.81e+00)= & 3.21e+01(3.38e+01)+ & \cellcolor[rgb]{0.8,0.8,0.8}{\textbf{1.03e+01(2.98e+00)}}  \\

F10 & 10 & 3.12e+01(2.58e+01)+ & 5.13e+01(2.48e+01)+ & \cellcolor[rgb]{0.8,0.8,0.8}{\textbf{1.20e+01(6.14e+00)}}  \\

F11 & 10 & 7.61e+00(2.66e+00)= & 9.49e+00(3.28e+00)= & \cellcolor[rgb]{0.8,0.8,0.8}{\textbf{7.54e+00(2.24e+00)}}  \\

F12 & 10 & \cellcolor[rgb]{0.8,0.8,0.8}{\textbf{9.96e+01(2.49e+02)}}= & 1.19e+04(3.04e+04)= & 5.70e+02(7.16e+02)  \\

F13 & 10 & 2.14e+00(1.23e+00)+ & 1.89e+02(5.14e+02)+ & \cellcolor[rgb]{0.8,0.8,0.8}{\textbf{1.32e+00(1.44e+00)}}  \\

F14 & 10 & 4.03e+00(2.95e-01)= & 4.07e+00(3.45e-01)= & \cellcolor[rgb]{0.8,0.8,0.8}{\textbf{4.00e+00(1.99e-01)}}  \\

F15 & 10 & 3.74e+02(1.08e+02)= & 5.29e+02(1.75e+02)= & \cellcolor[rgb]{0.8,0.8,0.8}{\textbf{3.48e+02(1.21e+02)}}  \\
\hline
+/=/- & & 7/8/0 & 8/7/0 & NA \\

Ranking & & 1.87 & 3.00 & \cellcolor[rgb]{0.8,0.8,0.8}{\textbf{1.13}} \\

$p$-value & & \cellcolor[rgb]{0.8,0.8,0.8}{\textbf{0.04}} & \cellcolor[rgb]{0.8,0.8,0.8}{\textbf{0}} & NA \\
\hline
\hline
\end{tabular}}
\label{tab:Statistics_Effextness_SRC_10D_results}
\end{table}
According to the results summarized in Table~\ref{tab:Statistics_Effextness_SRC_10D_results}, \textbf{LLM-SAEA} exhibits a clear advantage over V-WoSRC and V-SRC-Certain in 7 and 8 of the 15 benchmark cases, respectively, with no significant inferiority observed in any problem. Furthermore, the Friedman test indicates that \textbf{LLM-SAEA} attains the best average ranking and exhibits significant superiority over these two variants. Additionally, regarding the SRC, we offer the following analysis:
\begin{itemize}
    \item Compared to V-WoSRC, \textbf{LLM-SAEA} leverages the SRC to assign confidence labels to the output results, thereby enabling users to more effectively assess and utilize the information provided by the LLM.

    \item Compared to V-SRC-Certain, \textbf{LLM-SAEA} includes actions labeled as "certain" in the recommended action set $\mathcal{A}^*$ and also selects alternative actions to replace those with "uncertain" labels. Typically, actions in $\mathcal{CA}$ with higher average historical scores are prioritized. This approach effectively reduces errors in the LLM's reasoning processes by leveraging the historical information of each action.

\end{itemize}
\subsection{Parameter Sensitivity Analysis}
\textbf{LLM-SAEA} incorporates only one parameter: the population size $N$. We set $N=100$ to ensure consistency with most of the comparative algorithms. Additionally, we examine the impact of population size on the algorithm's performance by selecting five representative values: 60, 80, 100, 120, and 140. The experiment is carried out on the F1-F15 (\textbf{10D}) test problems, with the algorithm independently run 20 times for each value to attain statistical results. Table \ref{tab: parameter_N_analysis} presents the final statistical results. As evidenced by the results in Table \ref{tab: parameter_N_analysis}, selecting a smaller population size of $N=60$ significantly reduces the algorithm's performance. Lastly, we recommend a population size range of [80, 120].

\begin{table}[htbp]
\renewcommand{\arraystretch}{1.1}
\scriptsize

\caption{The ranking of the result on the F1-F15(\textbf{10D}) problems obtained by various $N$ values using the Friedman test with the Hommel post-hoc method applied to correct the resulting \emph{p}-value.}
\centering
\begin{tabular}{p{1.5cm}p{0.7cm}p{0.7cm}p{0.7cm}p{0.7cm}p{0.7cm}p{0.7cm}p{0cm}}
\hline
\hline
$N$	 &60	&80  &100 &120 &140 \\
\hline
Ranking   & 4.0 &2.7 &\cellcolor[rgb]{0.8,0.8,0.8}{\textbf{2.5}} &2.7 &3.5 \\
$p$-value &\cellcolor[rgb]{0.8,0.8,0.8}{\textbf{0.01}} &1  &NA  &1 &0.43 \\
\hline
\hline
\end{tabular}
\label{tab: parameter_N_analysis}
\end{table}

\section{Conclusion}
To enhance the efficiency and generalization capability of SAEAs, this paper proposes \textbf{LLM-SAEA}, a novel method based on a customized collaboration-of-experts framework that utilizes LLMs for the online automatic configuration of key components in SAEAs (e.g., surrogate models and infill sampling criteria). Specifically, one LLM acts as a decision expert responsible for selecting models and infill sampling criteria. At the same time, another LLM serves as a scoring expert, quantifying the utility of the adopted models and infill sampling criteria through numerical scores. Experimental results show that \textbf{LLM-SAEA} outperforms several state-of-the-art algorithms across a range of benchmark problems. Furthermore, a series of ablation studies validate the dynamic configuration capability of \textbf{LLM-SAEA}, the rationality of the collaboration-of-experts framework, and the effectiveness of the self-reflection component in \textbf{LLM-DE}.

In future research,  we aim to explore more advanced prompting techniques for LLMs to improve their performance in surrogate-assisted evolutionary optimization. Additionally, we plan to extend and validate the proposed algorithm to a broader range of real-world application problems. 
    



\section*{Disclosure of Interests} The authors have no competing interests to declare that are relevant to the content of this article.

\bibliographystyle{model1-num-names}

\bio{}
\endbio

\bio{}
\endbio

\end{CJK*}
\end{document}